%% file: acl_latex.tex
\documentclass[11pt]{article}

\usepackage[preprint]{acl}

\usepackage{times}
\usepackage{latexsym}

\usepackage[T1]{fontenc}
\usepackage[utf8]{inputenc}

\usepackage{microtype}

\usepackage{inconsolata}

\usepackage{graphicx}
\usepackage{xcolor} % 放在导言区

\usepackage{pifont} % provide \ding command
\definecolor{kellygreen}{rgb}{0.3, 0.73, 0.09}
\definecolor{alizarin}{rgb}{0.82, 0.1, 0.26}
\newcommand{\cmark}{{\color{kellygreen} \ding{51}}}
\newcommand{\xmark}{{\color{alizarin} \ding{55}}}

\definecolor{deepred}{rgb}{0.698,0.133,0.133}
\definecolor{blue}{rgb}{0,0,1}
\usepackage{tikz}
\usepackage{amssymb}
\newcommand\encircle[2][]{\tikz[overlay]\node[fill=blue!20,inner sep=2pt, anchor=text, rectangle, rounded corners=1.5mm,#1] {#2};\phantom{#2}}
\definecolor{lightcoral}{rgb}{0.94, 0.5, 0.5}
\definecolor{myblue}{rgb}{0.27,0.52,0.95}
\definecolor{harvestgold}{rgb}{0.85, 0.57, 0.0}

\usepackage{amsmath}
\usepackage{multirow}
\usepackage{subcaption}

\usepackage{xspace}
\usepackage[shortlabels]{enumitem}
\usepackage{booktabs}       % professional-quality tables

\usepackage{adjustbox}  % 加上这个
\usepackage{bbding}
\makeatletter
  \newcommand\figcaption{\def\@captype{figure}\caption}
  \newcommand\tabcaption{\def\@captype{table}\caption}
\makeatother

\usepackage{colortbl}  % 用于给表格单元格上色
\definecolor{wkred}{RGB}{255, 190, 190}
\definecolor{wkblue}{RGB}{210, 230, 250}
\definecolor{skyblue}{RGB}{117,167,211}  % 自定义天蓝色（DodgerBlue）

\usepackage[most]{tcolorbox}
\usepackage{xcolor}
\usepackage{cleveref}

\newtcolorbox{promptbox}[1]{
  colback=gray!5!white,
  colframe=gray!75!black, % 类似图片中的深灰色
  coltitle=white,
  fonttitle=\bfseries, % 标题字体变大
  title={#1},
  halign title=center, % 强制标题居中
  enhanced,
  arc=2mm, % 圆角
  attach boxed title to top center={yshift=-3mm},
  boxrule=1.5pt, % 边框粗细
  center, % 盒子整体居中
  left=2mm, right=2mm, top=3mm, bottom=2mm,
  fontupper=\small\ttfamily % 使用等宽字体模拟代码/提示词感
}

\definecolor{darkgreen}{RGB}{43, 195, 68} % 自定义一个暗绿色
\definecolor{lightgray}{gray}{0.9} % 0.9 是灰度值，越接近 1 越白

\title{Geoparsing: Diagram Parsing for Plane and Solid Geometry with a Unified Formal Language}

\author{Peijie Wang$^{1,2}$, Ming-Liang Zhang$^{3}$, Jun Cao$^{1,2}$, Chao Deng$^{1,2}$,  Dekang Ran$^{1,2}$, Hongda Sun$^{3}$\\ \textbf{Pi Bu$^{3}$, Xuan Zhang$^{3}$, Yingyao Wang$^{3}$, Jun Song$^{3}$, Bo Zheng$^{3}$, Fei Yin$^{1,2}$, Cheng-Lin Liu$^{1,2}$\thanks{~~~Corresponding authors.}}\\ 
  $^1$MAIS, Institute of Automation of Chinese Academy of Sciences\\
  $^2$School of Artificial Intelligence, University of Chinese Academy of Sciences\\
  $^3$Future Living Lab of Alibaba\vspace{0.0cm}\\
  \texttt{wangpeijie2023@ia.ac.cn} \quad \texttt{\{fyin, liucl\}@nlpr.ia.ac.cn}\\
  \texttt{\{zhangmingliang.zml, bupi.wj, jsong.sj\}@libaba-inc.com}\\ 
  \rule{0pt}{1.5cm} % 插入一个 0 宽度、1.5cm 高度的隐形支柱 (Strut)
}

\begin{document}
\maketitle
% \vspace*{10mm}

\input{sec/0_abstract}   
\input{sec/1_intro}

\input{sec/2_related_works}
\input{sec/3_language}
\input{sec/4_dataset}
\input{sec/5_method}
\input{sec/6_experiments}
\input{sec/7_conclusion}
\input{sec/limitations}

\bibliography{custom}

\input{appendix/X_suppl}

\end{document}

%% file: sec/0_abstract.tex
\begin{abstract}
\label{sec:abstract}
Multimodal Large Language Models (MLLMs) have achieved remarkable progress but continue to struggle with geometric reasoning, primarily due to the perception bottleneck regarding fine-grained visual elements. While formal languages have aided plane geometry understanding, solid geometry which requires spatial understanding remains largely unexplored. In this paper, we address this challenge by designing a unified formal language that integrates plane and solid geometry, comprehensively covering geometric structures and semantic relations. We construct GDP-29K, a large-scale dataset comprising 20k plane and 9k solid geometry samples collected from diverse real-world sources, each paired with its ground-truth formal description. To ensure syntactic correctness and geometric consistency, we propose a training paradigm that combines Supervised Fine-Tuning with Reinforcement Learning via Verifiable Rewards. Experiments show that our approach achieves state-of-the-art parsing performance. Furthermore, we demonstrate that our parsed formal descriptions serve as a critical cognitive scaffold, significantly boosting MLLMs' capabilities for downstream geometry reasoning tasks. Our data and code are available at \href{https://eternal8080.github.io/geoparsing.github.io/}{Geoparsing}.

\end{abstract}

%% file: sec/1_intro.tex
\section{Introduction}
\label{sec:intro}
Geometry plays a crucial role in mathematics and is widely considered its core~\cite{riemannian}; it has always been a subject of great interest in the field of artificial intelligence~\cite{alphageometry, geoeval}. The challenge of geometric problem solving lies in the integration of complex visual information and symbolic reasoning. Based on the structural properties of geometry diagram, geometry can be categorized into plane geometry and solid geometry~\cite{geoclass}. Compared to plane geometry, solid geometry demands understanding 3D structures and spatial relationships, making it more complex and a highly challenging problem for artificial intelligence systems~\cite{chou1996automated, mathvision}.

\begin{figure}[t]
    \centering
    \includegraphics[width=1.0\linewidth]{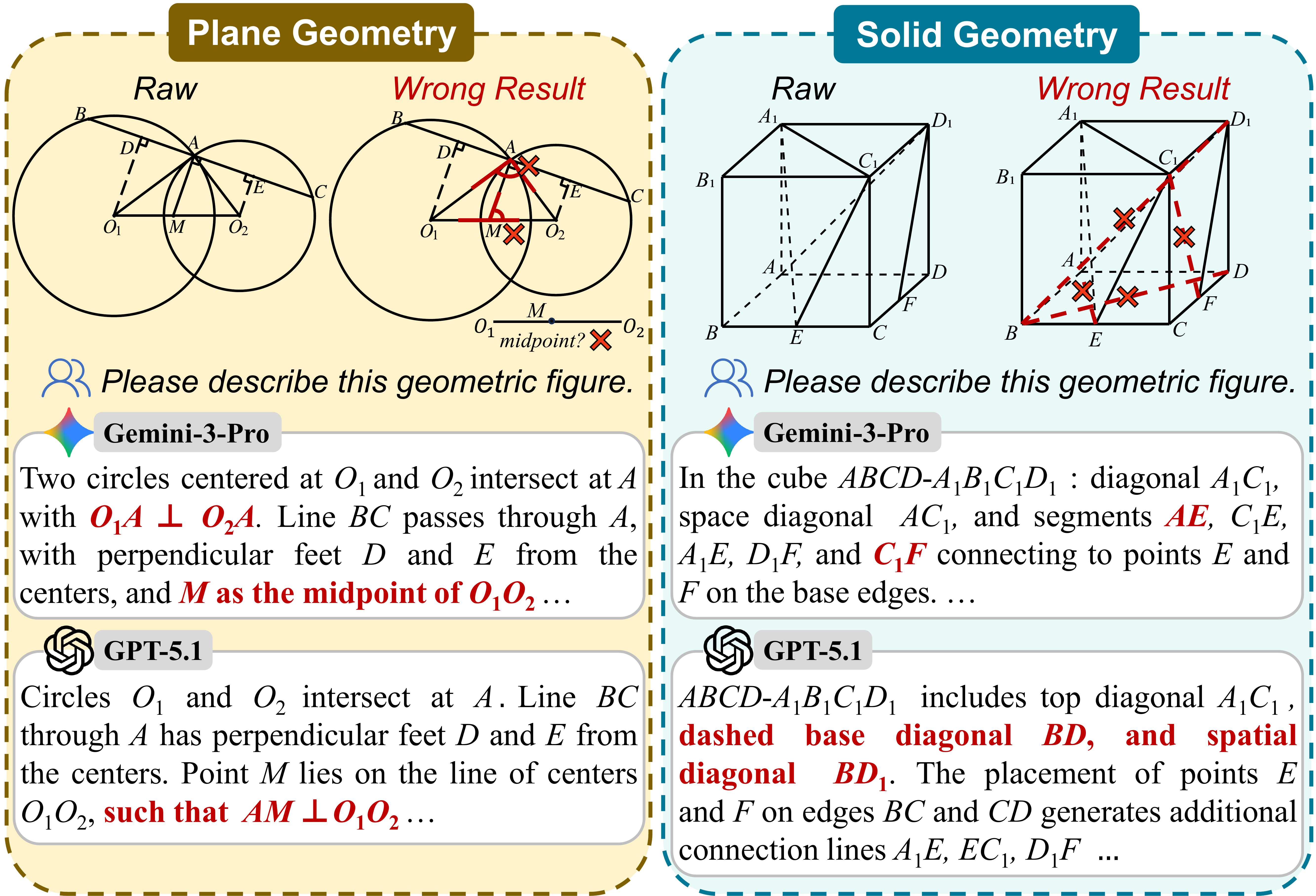}
    % \vspace{-5mm}
    % \caption{Hallucinations in geometric parsing by SOTA MLLMs. Gemini3-Pro and GPT-5.1 struggle to correctly parse slightly complex plane geometry and simple solid geometry. \textcolor{red}{Red} text indicates parsing errors.}
    \caption{Hallucinations in geometric parsing by SOTA MLLMs. Gemini-3-Pro and GPT-5.1 struggle to correctly parse slightly complex plane geometry and simple solid geometry. \textcolor{red!60!black}{Red} text indicates parsing errors.}
    % \caption{Geometry diagram parsing challenges for MLLMs. Even SOTA models (Gemini3-Pro, GPT-5.1) struggle to correctly parse slightly complex plane geometry and simple solid geometry. \textcolor{red}{Red} text indicates parsing errors, such as hallucinations of non-existent lines or misinterpretation of spatial structures.}
    \vspace{-5mm}
\label{fig:example}
\end{figure}

Recent advancements in Multimodal Large Language Models (MLLMs) have demonstrated remarkable capabilities across various vision reasoning tasks~\cite{llavaonevision,internvl3.5,Qwen3-VL,sun2024determlr,sun2024harnessing,deng2025longdocurl,kang2026vlm,lu2026mllms,zhang2025perl}. However, Geometry Problem Solving (GPS) remains a challenge~\cite{survey1,survey2}. The core difficulty stems from the strict demand for precise geometric perception: MLLMs must accurately identify basic geometric primitives (e.g., points, lines, and planes) and comprehend their relations. Yet, even state-of-the-art (SOTA) models frequently misinterpret geometric figures~\cite{solidgeo, eagle}. As shown in Figure~\ref{fig:example}, the most advanced models, Gemini-3-Pro~\cite{gemini3pro} and GPT-5.1~\cite{gpt5}, still struggle to correctly parse complex plane and solid geometries. This deficiency in fine-grained visual perception is a critical bottleneck, constraining the subsequent reasoning process.

\begin{figure*}[t]
    \centering
    \includegraphics[width=1.0\linewidth]{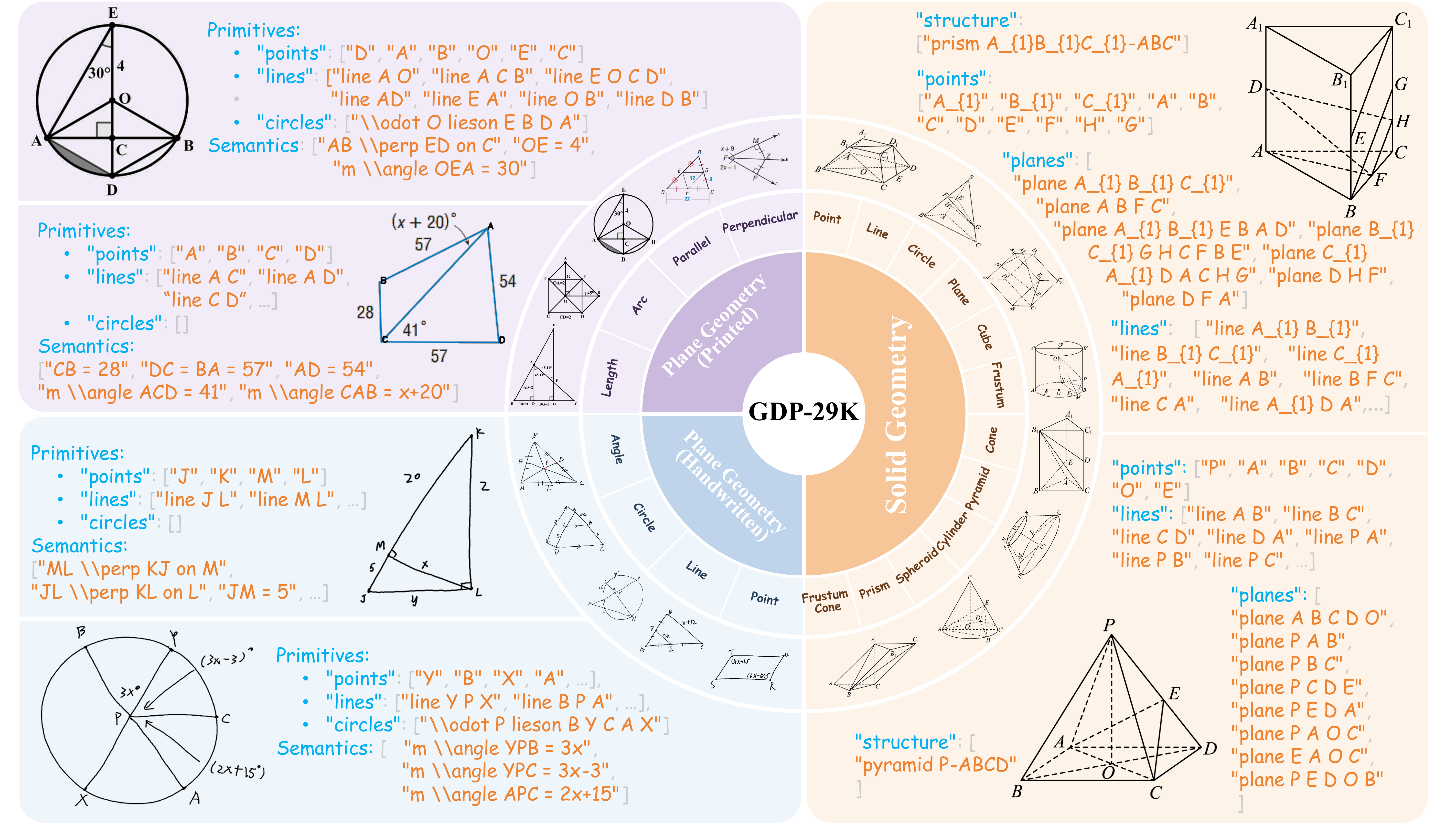}
    % \vspace{-5mm}
    \caption{Overview of the GDP-29K dataset for geometry diagram parsing. The dataset spans plane geometry (printed and handwritten) and solid geometry, with each diagram annotated by formal language that captures primitives and semantic constraints.}
    \vspace{-5mm}
\label{fig:overview}
\end{figure*}

To address the perception challenge, recent research has explored geometry diagram parsing (GDP), aiming to convert geometric diagrams into symbolic representations~\cite{seo2014diagram,intergps}. However, existing works predominantly focus on plane geometry (PGDP), introducing formal languages and datasets like PGDP5K~\cite{pgdpnet} and FormalGeo7K~\cite{formalgeo1}, while solid geometry remains underexplored. Unlike PGDP, solid geometry diagram parsing (SGDP) necessitates understanding of 3D spatial structures, making this task more complex~\cite{solidgeo}. To bridge this gap, we propose a unified formal language that extends established plane geometry formal representations to solid geometry. The formal language covers elements ranging from basic points and lines to high-order structures like planes and solids. Leveraging this language, we construct \textbf{GDP-29K}, a large-scale dataset sourced from diverse real-world scenarios. It comprises a plane geometry subset \textbf{PGDP-20K} and a solid geometry subset \textbf{SGDP-9K}, with each image paired with its ground-truth formal description. Notably, the dataset incorporates varied visual styles, including handwritten diagrams, significantly enriching data diversity. GDP-29K not only expands the scale of plane geometry resources but also fills the void in formal definitions and benchmarks for solid geometry.

Leveraging GDP-29K, we employ a two-stage training paradigm that integrates Supervised Fine-Tuning (SFT) with Reinforcement Learning via Verifiable Rewards (RLVR). To ensure the rigor of the generated formal descriptions, we design a rule-based verifier that guides the policy based on syntactic correctness and geometric consistency. Consequently, our model demonstrates superior parsing capabilities, with scores of 96.4 on PGDP and 94.9 on SGDP benchmark, even surpassing GPT-5.2 and Gemini-3-Flash. Furthermore, we demonstrate the practical utility of our parser in downstream geometry reasoning. Experimental results show that augmenting Qwen3-VL-8B with our parsing outputs drives significant performance boosts, yielding improvements of +10.1\% on Geometry3K~\cite{intergps}, +9.0\% on PGPS9K~\cite{pgps9k}, and +3.1\% on SolidGeo~\cite{solidgeo}, with gains also verified across other representative models. In summary, Our contributions are as follows:

\begin{itemize}[leftmargin=*]
    \item We propose a unified formal language for GDP task, which extends existing plane geometry representations to cover solid geometry structure.

    \item We construct \textbf{GDP-29K}, a large-scale dataset comprising 20K plane and 9K solid geometry diagrams paired with formal descriptions across both printed and handwritten styles, effectively filling the critical data gap for the GDP task.
    
    \item We introduce a robust training paradigm combining SFT and RLVR, which ensures syntactic and geometric validity while achieving SOTA performance on GDP benchmarks.

    \item Experimental results demonstrate that our parsing outputs significantly enhance downstream multimodal geometric reasoning.

\end{itemize}

%% file: sec/2_related_works.tex
\section{Related Work}
\label{sec:related_work}

\paragraph{Geometry Perception Limitations in MLLMs.}
Recent MLLMs have demonstrated strong capabilities on several mathematical reasoning benchmarks, such as MathVista~\cite{mathvista}, MathVision~\cite{mathvision}, and We-Math~\cite{wemath}. However, performance remains unsatisfactory in GPS~\cite{geosense, solidgeo}. Solving geometry problems requires the model to accurately identify fundamental primitives such as points, lines, circles, and planes; failure to perceive these elements correctly inevitably leads to reasoning errors. Most existing works on GPS rely on end-to-end benchmarks like GeoEval~\cite{geoeval} and MathVerse~\cite{mathverse}. However, this approach tends to conflate perception errors with reasoning failures, obscuring the true source of model limitations. In fact, several studies have identified that perception errors remain the primary source of failure in geometric reasoning tasks~\cite{mvmath, solidgeo}. Thus, explicitly decoupling perception from reasoning is imperative.

\paragraph{GDP Datasets and Formalization.}
GDP translates geometric diagrams into formal languages to decouple perception from reasoning. While PGDP benchmarks like Geometry3K~\cite{intergps}, PGDP5K~\cite{pgdpnet}, and FormalGeo7k~\cite{formalgeo1} have established 2D formalisms, their reliance on limited sources (e.g., Geometry3K and GeoQA~\cite{geoqa}) restricts visual and structural diversity. This lack of variety potentially hinders the generalization of parsing models across complex, real-world scenarios. Critically, a significant gap remains in solid geometry, which involves complex 3D structures and spatial relationships~\cite{solidgeo} unaddressed by current formalisms and datasets. To bridge this, we design a formal language for solid geometry that is fully compatible with 2D representations, and introduce GDP-29K—a large-scale dataset comprising 9K solid and 20K plane geometry samples. This resource fills the long-standing void in solid geometry parsing while significantly enhancing the diversity and scale of plane geometry benchmarks.

\paragraph{Approaches for Geometry Understanding and Reasoning.}
Early approaches relied on rule-based heuristics~\cite{intergps} or detection-based pipelines~\cite{pgdpnet} to identify geometric primitives. Recently, works like G-LLaVA~\cite{gllava}, AutoGeo~\cite{autogeo}, and MAVIS~\cite{mavis} have shifted the focus toward geometry QA, utilizing natural language supervision to enhance reasoning. While GeoX~\cite{geox} validates the feasibility of formal language pre-training, many current methods still struggle with the structural rigor required for precise parsing. Inspired by the success of Reinforcement Learning (RL) in mathematical domains~\cite{deepseekmath, deepseekr1}, we introduce RLVR to the GDP task—marking the first application of RL to ensure both syntactic correctness and geometric precision in diagram parsing.

%% file: sec/3_language.tex
\section{Geometry Formal Representation}
\label{sec:representation}

In this section, we introduce our unified formal language representation. Designed for conciseness and compatibility, this framework extends existing definitions to address the critical lack of formalisms for solid geometry.

\paragraph{Inheritance from Plane Geometry.}
For plane geometry, we adopt the formal language established in PGPS9K~\cite{pgps9k}. This representation is concise and close to natural language, describing geometric diagrams as sequences of predicates. It covers fundamental primitives (e.g., points, lines, circles) and semantic relations, including geometric constraints (e.g., parallelism, perpendicularity) and metric attributes (e.g., lengths, angle measures), effectively capturing the topological structure of plane geometry.

\paragraph{Extension to Solid Geometry.}
To address the lack of formal definitions for three-dimensional structures, we extend the formal language to solid geometry. While preserving the syntactic consistency of the plane geometry language, we introduce high-order primitives such as \textit{planes} and \textit{solids}. To achieve comprehensive coverage, we explicitly categorize solid structures into two classes:

\begin{itemize}[leftmargin=*]
    \item \textbf{Polyhedra:} We design specific descriptors for a wide array of multifaceted bodies, ranging from basic forms like \textit{cubes}, \textit{prisms}, and \textit{pyramids} to more complex structures such as \textit{frustums} and \textit{composite polyhedra}.
    \item \textbf{Solids of Revolution:} We strictly define curved geometric bodies formed by rotating a plane curve around an axis, including \textit{spheres}, \textit{cylinders}, \textit{cones}, and \textit{truncated cones}.
\end{itemize}

For each category, we establish standardized formal templates to ensure structural consistency across diverse solid configurations. This language is characterized by its modularity and high expressiveness, allowing intricate geometric structures to be decomposed into interpretable primitives. Such a design ensures full compatibility with existing plane geometry datasets while enabling the precise description of complex solid structures.

\begin{table}[t]
    \centering
    \small
    % 增加行间距
    \renewcommand{\arraystretch}{1.1} 
    % 调整列间距
    \setlength{\tabcolsep}{10pt} 
    
    \begin{tabular}{lr}
    \toprule
    \textbf{Statistic} & \textbf{Number} \\
    \midrule
    \multicolumn{2}{l}{\textit{\textbf{Dataset Scale \& Style}}} \\
    Total Images & 28,882 \\
    \hspace{3mm}- Plane Geometry (PG) & 19,965 \\
    \hspace{3mm}- Solid Geometry (SG) & 8,917 \\
    Style &  \\
    \hspace{3mm}- Printed & 23,366 \\
    \hspace{3mm}- Handwritten & 5,516 \\
    \midrule
    \multicolumn{2}{l}{\textit{\textbf{PG Statistics (Avg. per Image)}}} \\
    Points & 5.9 \\
    Lines & 5.0 \\
    Circles & 0.3 \\
    Semantic Relations & 2.4 \\
    \midrule
    \multicolumn{2}{l}{\textit{\textbf{SG Statistics (Avg. per Image)}}} \\
    Points & 7.4 \\
    Lines & 12.1 \\
    Circles & 0.05 \\
    Planes & 6.4 \\
    \bottomrule
    \end{tabular}
    
    \caption{Detailed statistics of GDP-29K.}
    \label{tab:statistics}
    \vspace{-5mm}
\end{table}

%% file: sec/4_dataset.tex
\section{GDP-29K Dataset}
\label{sec:dataset}

\subsection{Overview}
\label{subsec:overview}
Based on the formal language defined in Section~\ref{sec:representation}, we construct \textbf{GDP-29K}, a large-scale dataset designed to advance geometric diagram parsing tasks. GDP-29K comprises a total of 28,977 samples collected from diverse real-world scenarios, with each image paired with its ground-truth formal description. The dataset contains two subsets:
\begin{itemize}[leftmargin=*]
    \item \textbf{PGDP-20K:} Containing 19,965 plane geometry diagrams. PGDP-20K incorporates a wide spectrum of visual styles, covering both printed diagrams and handwritten sketches. This variety significantly enriches data diversity, enhancing the robustness of model training.
    \item \textbf{SGDP-9K:} Containing 8,917 solid geometry diagrams. To the best of our knowledge, this constitutes the first large-scale dataset tailored for solid geometry parsing, effectively filling the gap in data resources for 3D geometry perception.
\end{itemize}

Figure~\ref{fig:overview} illustrates representative samples from both subsets, highlighting the complexity and diversity of the geometric structures. Detailed statistics of GDP-29K are presented in Table~\ref{tab:statistics}.

\subsection{Dataset Construction}
\label{subsec:construction}
The construction of GDP-29K follows a rigorous pipeline comprising data collection, filtering, and labeling, ensuring both diversity and high quality.
\paragraph{Data Collection.}
We aggregated raw geometric images from a wide range of real-world sources, including open-access textbooks, exam papers, and educational websites\footnote{\url{https://www.jiaoyanyun.com/}}. To further enhance diversity, we also curated samples from three existing open-source datasets~\cite{pgdpnet,codeplot,auxsolidmath}. In this initial phase, we accumulated a raw pool of 68,642 plane geometry images and 28,878 solid geometry images.

\begin{table*}[t]
    \centering
    \resizebox{1.0\textwidth}{!}{
    \begin{tabular}{l|cccc|ccc|cc}
        \toprule
        
        \textbf{Benchmarks}& Language & PG Size & SG Size & Task 
        & Style & Source & SG category & PGFL & SGFL \\
        \midrule
        GeoQA~\cite{geoqa}& CN & 4849 & 115& GQA & \encircle[fill=pink, text=white]{P}  & \encircle[fill=harvestgold, text=white]{S}  & 4 & \xmark & \xmark \\

        Geometry3K~\cite{intergps}& EN & 3002 & 0 & GQA & \encircle[fill=pink, text=white]{P}  & \encircle[fill=harvestgold, text=white]{S}  & \xmark & \cmark & \xmark \\

        PGDP5K~\cite{pgdpnet}& EN & 5000 & 0 & PGDP & \encircle[fill=pink, text=white]{P}  & \encircle[fill=harvestgold, text=white]{S}  & \xmark & \cmark & \xmark \\

        Formalgeo7k~\cite{formalgeo2}& EN & 7000 & 0 & MQA & \encircle[fill=pink, text=white]{P}  & \encircle[fill=harvestgold, text=white]{O}  & \xmark & \cmark & \xmark \\
        
        GeoEval~\cite{geoeval} & EN & 2000 & 272 & GQA & \encircle[fill=pink, text=white]{P} & \encircle[text=white]{O} & 3 & \xmark & \xmark \\
        
        MATH-Vision~\cite{mathvision} & EN & 1122& 244 & MQA & \encircle[fill=pink, text=white]{P} & \encircle[fill=harvestgold, text=white]{S} & 4 & \xmark & \xmark \\

        OlympiadBench~\cite{olympiadbench} & EN/CN& 1325 & 1322 & MQA & \encircle[fill=pink, text=white]{P} & \encircle[fill=harvestgold, text=white]{S} & 6 & \xmark & \xmark \\
        
        MathVerse~\cite{mathverse}& EN & 1746 & 332 & MQA & \encircle[fill=pink, text=white]{P} & \encircle[fill=harvestgold, text=white]{S} \encircle[text=white]{O} & 4  & \xmark & \xmark \\

        MV-MATH~\cite{mvmath}& EN & 1175& 372 & MQA & \encircle[fill=pink, text=white]{P}  & \encircle[fill=harvestgold, text=white]{S}  & 6 & \xmark & \xmark \\
        
        GeoSense~\cite{geosense}& EN/CN & 1558 & 231 & GQA & \encircle[fill=pink, text=white]{P}  & \encircle[fill=harvestgold, text=white]{S} \encircle[text=white]{O}  & 6 & \xmark & \xmark \\

       SolidGeo~\cite{solidgeo} & EN/CN& 0& 3113 & GQA & \encircle[fill=pink, text=white]{P} & \encircle[fill=harvestgold, text=white]{S} \encircle[text=white]{O} & 9 & \xmark & \xmark \\
        
        \midrule
        
        GDP-29K(Ours) & EN & 19965 & 8917 & GDP& \encircle[fill=pink, text=white]{P}  \encircle[fill=orange, text=white]{H} & \encircle[fill=harvestgold, text=white]{S} \encircle[text=white]{O} & 9 & \cmark & \cmark \\
        \bottomrule
    \end{tabular}}
    \caption{Comparison with existing multimodal math benchmarks. SG: Solid Geometry, PG: Plane Geometry. GQA: Geometry QA, MQA: Math QA.\textbf{Style:} $\encircle[fill=pink, text=white]{P}$=\underline{P}rinted, 
    $\encircle[fill=orange, text=white]{H}$=\underline{H}andwritten. \textbf{Source:} $\encircle[fill=harvestgold, text=white]{S}$=\underline{S}elf-sourced, $\encircle[text=white]{O}$=Collected from \underline{O}pen-source Dataset.
    PGFL/SGFL: Plane/Solid Geometry Formal Language.}
    \label{tab:comparison}
    \vspace{-2mm}
\end{table*}

\paragraph{Data Filtering.}
To ensure the quality of the collected data, we use a three-stage filtering pipeline:
\begin{itemize}[leftmargin=*]
    % \vspace{-3mm}
    \item Stage 1: Image Quality Filtering. Using OpenCV, we computed sharpness metrics to eliminate samples with low-resolution or blurry images, ensuring the retained diagrams were visually clear.
    % \vspace{-5mm}
    \item Stage 2: Semantic Quality Filtering. Leveraging GPT-5.1's visual understanding, we filtered out images with semantic ambiguity, unsuitable for parsing, or unrecognizable text labels.
    % \vspace{-3mm}
    \item Stage 3: Human Verification. Finally, we conducted a comprehensive manual review of retained images, strictly excluding poorly rendered diagrams or severe artifacts hindering parsing.
    % \vspace{-3mm}
\end{itemize}

After this rigorous filtering process, we obtained a refined set of 22,459 plane geometry samples and 9,541 solid geometry samples.

\paragraph{Data Labeling.} We adopted different annotation strategies for plane and solid geometry. For the former, we utilized a model-assisted pipeline where GPT-5.1 generated initial formal descriptions, which were subsequently refined by expert annotators to accelerate the process. In contrast, solid geometry requires a purely manual approach from scratch to ensure structural rigor, as current MLLMs still struggle with 3D spatial perception. To guarantee the highest label quality, we implemented a strict three-tier quality control protocol—consisting of \textit{Annotation}, \textit{Verification}, and \textit{Final Acceptance}—ensuring that only samples passing all stages were included in the final dataset. Following the annotation, we performed a final redundancy filtering step by identifying and removing samples with identical formal descriptions. This ensured the structural uniqueness of each instance, resulting in the final 28,977 high-quality samples.

\subsection{Comparison with Existing Benchmarks}
As shown in Table~\ref{tab:comparison}, GDP-29K advances geometry diagram parsing in two key dimensions. First, in plane geometry, its 19,965 real-world samples surpass the cumulative scale of major existing benchmarks, such as Geometry3K~\cite{intergps} and PGDP5K~\cite{pgdpnet}, enabling more robust 2D geometric perception. Second, GDP-29K introduces the first formal representation and dataset for solid geometry, featuring 8,917 samples. This fills a critical void in a domain essential for geometric reasoning that previous works have entirely neglected. 

%% file: sec/5_method.tex
\section{Methodology}
\label{sec:method}

\begin{figure*}[t]
    \centering
    \includegraphics[width=\textwidth]{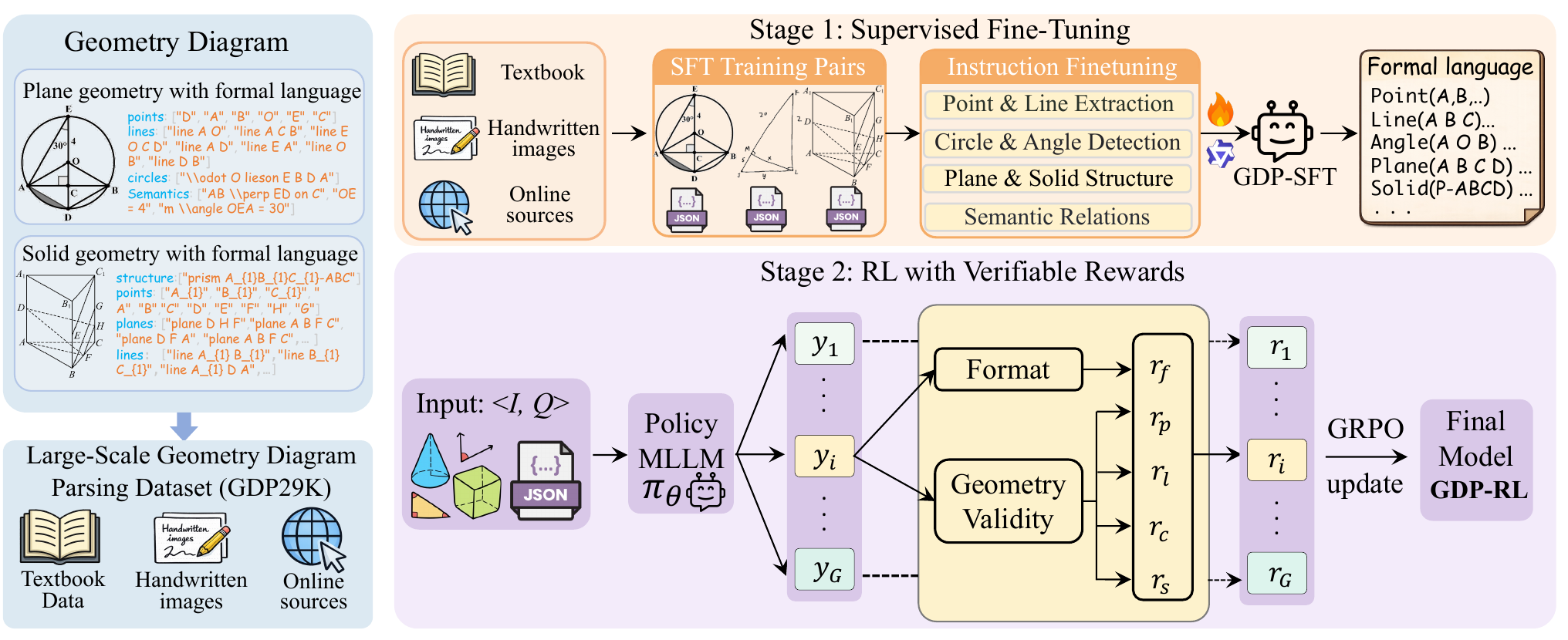}
    \caption{Overview of our geometry diagram parsing framework. We first construct SFT training pairs from GDP-29K and obtain initial parser GDP-SFT via instruction fine-tuning. We then further optimize the parser with reinforcement learning via verifiable rewards that enforce both format correctness and geometric validity. The final model GDP-RL generates unified formal language descriptions for both plane and solid geometry diagrams.}
    \label{fig:framework}
\end{figure*}

Our goal is to develop a multimodal model $\mathcal{M}_\theta$ that, given a visual geometry diagram $I$ and a parsing instruction $Q$, generates a formally rigorous description sequence $Y$. $Y$ consists of a set of geometry primitives and relations defined in our formal language. Formally, the model predicts:
\begin{equation}
    Y = \mathcal{M}_\theta(I, Q)
\end{equation}
To achieve this, we introduce a training framework comprising two stages: Supervised Fine-Tuning (SFT) for syntax alignment and Reinforcement Learning via Verifiable Rewards (RLVR) for enforcing syntactic rigor and geometric consistency. The training pipeline is shown in Figure~\ref{fig:framework}.

\subsection{Stage 1: Supervised Fine-Tuning}
\label{subsec:sft}
The initial stage aims to teach the base model the fundamental syntax of our formal language and the mapping between visual features and geometry primitives. Given the training dataset $\mathcal{D} = \{(I_i, Q_i, Y^{ref}_i)\}_{i=1}^N$, where $Y^{ref}$ denotes the ground-truth formal description, the objective is to maximize the likelihood of generating the correct sequence. The model $\mathcal{M}_\theta$ is fine-tuned by minimizing the cross-entropy loss:
\begin{equation}
    \mathcal{L}_{\text{SFT}}(\theta) = - \frac{1}{N} \sum_{i=1}^{N} \sum_{t=1}^{T} \log P_\theta(y_{i,t} \mid y_{i, <t}, I_i, Q_i)
\end{equation}
where $y_{i,t}$ is the $t$-th token of the sequence $Y_i$. This step uses standard teacher-forcing to ground the model in the correct formal syntax.

\begin{table*}[t]
    \centering    
    \resizebox{\textwidth}{!}{
    \begin{tabular}{l|ccc|ccc|ccc|ccc|c}
        \toprule
        \multirow{2}{*}{\textbf{Model}} & \multicolumn{3}{c|}{\textbf{Points}} & \multicolumn{3}{c|}{\textbf{Lines}} & \multicolumn{3}{c|}{\textbf{Circles}} & \multicolumn{3}{c|}{\textbf{Semantics}} & \multirow{2}{*}{\textbf{Overall}} \\
        \cmidrule(lr){2-4} \cmidrule(lr){5-7} \cmidrule(lr){8-10} \cmidrule(lr){11-13}
         & \textbf{P} & \textbf{R} & \textbf{F1} & \textbf{P} & \textbf{R} & \textbf{F1} & \textbf{P} & \textbf{R} & \textbf{F1} & \textbf{P} & \textbf{R} & \textbf{F1} & \textbf{Score} \\
        \midrule
        
        % --- Category 1: Traditional Methods ---
        \multicolumn{14}{c}{\cellcolor{lightgray}\textbf{\textit{Traditional Methods}}} \\
        InterGPS & 47.3 & 90.0 & 62.2 & 3.6 & 52.8 & 6.7 & 1.0 & 7.7 & 1.8 & 11.4 & 11.8 & 11.6 & 20.6 \\
        PGDPNet & 88.8 & 94.1 & 91.4 & 66.8 & 73.4 & 70.1 & 61.6 & 61.2 & 61.4 & 64.1 & 51.5 & 57.1 & 70.6 \\
        \midrule
        
        % --- Category 2: Open-source MLLMs ---
        \multicolumn{14}{c}{\cellcolor{lightgray}\textbf{\textit{Open-source MLLMs}}} \\
        LLaVA-OneVision-1.5-7B & 94.6 & 95.2 & 94.3 & 51.9 & 52.1 & 52.0 & 54.4 & 57.1 & 55.7 & 57.5 & 58.4 & 57.9 & 65.0 \\
        InternVL3.5-8B-Instruct & 96.2 & 94.8 & 95.6 & 61.9 & 58.7 & 60.2 & 62.3 & 87.5 & 72.8 & 63.1 & 59.4 & 61.2 & 72.5 \\
        Qwen3-VL-4B-Instruct & 95.8 & 97.3 & 96.6 & 53.5 & 71.7 & 61.2 & 75.5 & 92.7 & 83.3 & 60.5 & 57.5 & 58.9 & 75.0 \\
        Qwen3-VL-8B-Instruct & 97.1 & 95.5 & 96.3 & 56.4 & 69.8 & 62.4 & 76.8 & 94.3 & 84.6 & 66.2 & 55.4 & 60.1 & 75.9 \\
        Qwen3-VL-32B-Instruct & 98.4 & 96.5 & 97.4 & 72.9 & 75.2 & 74.0 & 86.2 & 93.5 & 89.6 & 61.8 & 60.7 & 61.3 & 80.5 \\
        Qwen3-VL-235B-Instruct & 99.0 & 96.6 & 97.8 & 84.5 & 80.3 & 82.3 & 85.7 & 93.8 & 89.5 & 65.2 & 69.7 & 67.4 & 84.2 \\
        Qwen3-VL-235B-Thinking & 99.0 & 96.6 & 97.8 & 91.7 & 89.1 & 90.4 & 89.7 & 92.3 & 91.0 & 76.4 & 68.8 & 72.4 & 87.9 \\
        \midrule
        
        % --- Category 3: Closed-source MLLMs ---
        \multicolumn{14}{c}{\cellcolor{lightgray}\textbf{\textit{Closed-source MLLMs}}} \\
        Claude-4.5-Sonnet & 95.9 & 92.7 & 94.3 & 72.7 & 73.4 & 73.0 & 89.6 & 87.5 & 88.5 & 68.3 & 70.1 & 69.1 & 81.2 \\
        GPT-5.2-1211 & 99.0 & 95.3 & 97.1 & 89.5 & 81.5 & 85.3 & 94.5 & 88.7 & 91.5 & 78.6 & 73.8 & 76.1 & 87.5 \\
        Gemini-3-Flash & 99.6 & 98.1 & 98.9 & \textbf{98.2} & 96.8 & 97.5 & 97.4 & 95.1 & 96.2 & 83.5 & 81.8 & 82.7 & 93.8 \\
        \midrule
        
        % --- Category 4: Ours ---
        \multicolumn{14}{c}{\cellcolor{lightgray}\textbf{\textit{Ours}}} \\
        \textbf{GDP-4B-SFT} & \textbf{99.7} & \textbf{99.6} & \textbf{99.6} & 96.8 & 97.7 & 97.3 & 97.4 & 97.3 & 97.4 & 87.8 & 87.3 & 87.5 & 95.1 \\
        \textbf{GDP-4B-RL} & 99.6 & 99.6 & 99.6 & 98.1 & \textbf{97.9} & \textbf{98.0} & \textbf{98.3} & \textbf{98.4} & \textbf{98.3} & \textbf{91.1} & \textbf{90.4} & \textbf{90.7} & \textbf{96.4} \\

        \bottomrule
    \end{tabular}
    }
    \caption{Performance comparison on the PGDP-2K test benchmark. Models are categorized by type. \textbf{P}: Precision, \textbf{R}: Recall, \textbf{F1}: F1-Score. \textbf{Semantics}: Semantic Relations. Overall Score represents the aggregate parsing accuracy. \textbf{Bold} denotes the best performance.}
    \label{tab:main_results}
    \vspace{-2mm}
\end{table*}

\subsection{Stage 2: RL with Verifiable Rewards}
\label{subsec:rlvr}
While SFT provides a strong foundation, it optimizes token-level likelihood rather than global structural integrity. Consequently, the model may generate outputs that are syntactically plausible but geometrically invalid. To address this, the second stage refines the policy $\pi_\theta$ using RLVR. The objective is to maximize the expected reward:
\begin{equation}
    \mathcal{J}(\theta) = \mathbb{E}_{(I, Q) \sim \mathcal{D}} \left[ \mathbb{E}_{Y \sim \pi_\theta(\cdot \mid I, Q)} \left[ R(Y) \right] \right]
\end{equation}
where $R(Y)$ is a scalar reward provided by our rule-based verifier. We optimize this objective using GRPO~\cite{deepseekmath}, which stabilizes training by normalizing rewards within sampled groups.

\paragraph{Verification Reward.}
We design a rule-based reward function $R(Y)$ to enforce both format compliance and semantic accuracy. The total reward is a weighted sum of two components:
\begin{equation}
    R(Y) = \lambda_1 R_{fmt}(Y) + \lambda_2 R_{geo}(Y)
\end{equation}
where $\lambda_1$ and $\lambda_2$ are hyperparameters balancing structural completeness and accuracy.

\paragraph{Format Reward ($R_{fmt}$).} 
To ensure the output adheres to the required structure, $R_{fmt}$ verifies the presence and correctness of tags (e.g., \texttt{<points>}, \dots, \texttt{<solids>}). It returns a binary signal $\mathbb{I}(Y \in \mathcal{F})$, where $\mathcal{F}$ denotes the set of sequences conforming to the predefined structural format.

\paragraph{Geometric Validity Reward ($R_{geo}$).} 
This component evaluates the alignment between the parsed primitives $\mathcal{P}$ and the ground-truth $\mathcal{P}^{\text{ref}}$. Recognizing that the $K$ types of primitives defined in our formal language (e.g., points, lines, planes, and semantic relations) present varying levels of perceptual difficulty, we implement a type-aware weighting strategy. For each primitive type $k \in \{1, \dots, K\}$, we assign a specific weight $\omega_k$ to reflect its complexity. The reward is computed as a weighted sum of the precision within each type:
\begin{equation}
    R_{geo}(Y) = \sum_{k=1}^{K} \omega_k \cdot \frac{|\mathcal{P}_k \cap \mathcal{P}_k^{\text{ref}}|}{|\mathcal{P}_k|}
\end{equation}
where $\mathcal{P}_k$ and $\mathcal{P}_k^{\text{ref}}$ denote the predicted and ground-truth subsets belonging to the $k$-th type, respectively. This granular reward structure encourages the model to maintain high fidelity across all geometric elements, especially for challenging high-order relations.

\begin{table*}[t]
    \centering
    
    % 稍微增加列间距，因为列数减少了
    \setlength{\tabcolsep}{4pt}
    
    \resizebox{\textwidth}{!}{
    \begin{tabular}{l|ccc|ccc|ccc|ccc|c|c}
        \toprule
        \multirow{2}{*}{\textbf{Model}} & \multicolumn{3}{c|}{\textbf{Points}} & \multicolumn{3}{c|}{\textbf{Lines}} & \multicolumn{3}{c|}{\textbf{Circles}} & \multicolumn{3}{c|}{\textbf{Planes}} & \textbf{Solids} & \textbf{Overall} \\
        \cmidrule(lr){2-4} \cmidrule(lr){5-7} \cmidrule(lr){8-10} \cmidrule(lr){11-13}
         & \textbf{P} & \textbf{R} & \textbf{F1} & \textbf{P} & \textbf{R} & \textbf{F1} & \textbf{P} & \textbf{R} & \textbf{F1} & \textbf{P} & \textbf{R} & \textbf{F1} & \textbf{Acc} & \textbf{Score} \\
        \midrule
    
        % --- Open-source ---
        \multicolumn{15}{c}{\cellcolor{lightgray}\textbf{\textit{Open-source MLLMs}}} \\
        LLaVA-OneVision-1.5-7B & 91.6 & 96.9 & 94.2 & 54.8 & 66.5 & 60.8 & 34.5 & 57.4 & 43.1 & 31.2 & 62.3 & 41.5 & 69.6 & 61.9 \\

        Qwen3-VL-4B-Instruct & 99.0 & 98.5 & 98.8 & 48.7 & 69.5 & 57.3 & 44.7 & 47.2 & 46.0 & 35.0 & 73.9 & 47.5 & 70.8 & 64.0 \\
        
        InternVL3.5-8B-Instruct & 98.5 & 94.8 & 96.6 & 57.8 & 50.4 & 53.8 & 23.3 & 86.1 & 36.7 & 62.4 & 72.6 & 67.1 & 69.7 & 64.8 \\
        
        Qwen3-VL-8B-Instruct & 98.9 & 99.1 & 99.0 & 42.3 & 65.4 & 51.4 & 48.9 & 57.6 & 52.9 & 32.0 & 75.1 & 44.8 & 77.5 & 65.1 \\
        
        Qwen3-VL-32B-Instruct & 99.0 & 99.1 & 99.0 & 63.6 & 77.9 & 70.0 & 42.9 & 91.1 & 58.4 & 27.5 & 82.1 & 41.2 & 83.8 & 70.4 \\

        Qwen3-VL-235B-Thinking & 99.2 & 95.2 & 97.1 & 86.7 & 75.3 & 80.6 & 33.3 & 77.2 & 46.6 & 94.5 & 45.1 & 61.0 & 83.2 & 73.7 \\

        Qwen3-VL-235B-Instruct & 98.7 & 93.9 & 96.2 & 75.3 & 73.2 & 74.2 & 51.5 & 83.1 & 63.6 & 72.6 & 75.5 & 74.0 & 83.0 & 78.2 \\
        \midrule
        % --- Closed-source ---
        \multicolumn{15}{c}{\cellcolor{lightgray}\textbf{\textit{Closed-source MLLMs}}} \\
        Claude-4.5-Sonnet & 95.7 & 96.1 & 95.9 & 70.1 & 72.6 & 71.8 & 28.5 & 85.2 & 42.7 & 64.8 & 73.4 & 68.8 & 78.3 & 71.5 \\
        
        GPT-5.2-1211 & 98.9 & 98.5 & 98.7 & 78.2 & 68.1 & 72.8 & 57.0 & 76.2 & 65.3 & 81.2 & 71.3 & 75.9 & 72.7 & 77.1 \\
        
        Gemini-3-Flash & \textbf{99.4} & \textbf{99.7} & \textbf{99.6} & \textbf{96.3} & 96.0 & 96.1 & \textbf{88.7} & 70.3 & 78.5 & 91.9 & 92.8 & 92.4 & 88.9 & 91.1 \\
        \midrule
        
        % --- Ours ---
        \multicolumn{15}{c}{\cellcolor{lightgray}\textbf{\textit{Ours}}} \\
        \textbf{GDP-4B-SFT} & 98.9 & 99.2 & 99.0 & 95.7 & 95.9 & 95.8 & 83.3 & 84.2 & 83.7 & 93.5 & 94.5 & 94.0 & 96.6 & 93.8 \\
        \textbf{GDP-4B-RL} & 99.2 & 99.3 & 99.2 & 96.1 & \textbf{96.8} & \textbf{96.4} & 86.7 & \textbf{86.2} & \textbf{86.5} & \textbf{95.8} & \textbf{95.2} & \textbf{95.5} & \textbf{97.0} & \textbf{94.9} \\
        
        \bottomrule
    \end{tabular}
    }
    \caption{Performance comparison on the SGDP-1K test benchmark. We report Precision (\textbf{P}), Recall (\textbf{R}), and F1-Score (\textbf{F1}) for multi-component primitives, and Accuracy (\textbf{Acc}) for the overall solid type. \textbf{Overall Score} represents the aggregate parsing accuracy. \textbf{Bold} denotes the best performance.}
    \label{tab:solid_results}
    \vspace{-0mm}
\end{table*}

\begin{table*}[t]
\centering
\small
\setlength{\tabcolsep}{6pt}
\resizebox{\textwidth}{!}{
\begin{tabular}{lccccc|cccccc}
\toprule
\multirow{2}{*}{Model} & \multicolumn{5}{c|}{\textbf{PGDP-2K}} & \multicolumn{6}{c}{\textbf{SGDP-1K}} \\
\cmidrule(lr){2-6} \cmidrule(lr){7-12}
& Points & Lines & Circles & Semantics & PPR
& Points & Lines & Circles & Planes & Solids & PPR \\
\midrule
Qwen3-VL-4B  & 81.5 & 33.4 & 73.8 & 37.3 & 27.4 & 93.1 & 38.6 & 42.6 & 32.8 & 70.8 & 26.2 \\
Qwen3-VL-8B  & 81.5 & 35.7 & 77.2 & 45.7 & 30.4 & 94.1 & 36.1 & 49.5 & 32.3 & 77.5 & 26.4 \\
Qwen3-VL-32B & 83.3 & 54.6 & 80.3 & 50.7 & 44.8 & 94.7 & 51.5 & 55.4 & 31.7 & 83.8 & 28.8 \\
GPT-5.2-1211          & 82.9 & 64.4 & 83.7 & 60.9 & 55.8 & 91.5 & 54.9 & 61.7 & 58.6 & 72.7 & 50.2 \\
Gemini-3-Flash        & 91.4 & 87.0 & 91.7 & 69.8 & 63.9 & \textbf{97.7} & 80.4 & 76.3 & 71.7 & 88.9 & 64.7 \\
\midrule
\textbf{GDP-4B-RL}    & 96.3 & 87.9 & 93.0 & 78.9 & 72.8 & 94.0 & 82.4 & 78.4 & 80.2 & 97.0 & 70.9 \\
\bottomrule
\end{tabular}
}
\caption{Sample Accuracy (SA) and Perfect Parsing Rate (PPR) on PGDP-2K and SGDP-1K.}
\label{tab:sa_ppr_combined}
\end{table*}

%% file: sec/6_experiments.tex
\section{Experiments}

\subsection{Experimental Setup}
\label{subsec:setup}

\paragraph{Datasets and Metrics.}
We conduct experiments on the GDP-29K dataset. We split it into a training set GDP-26K and a test benchmark GDP-3K. The GDP-3K test set is further divided into a plane-geometry subset PGDP-2K and a solid-geometry subset SGDP-1K. We report Precision (P), Recall (R), and F1-score (F1) for each primitive category by comparing the predicted and ground-truth sets.

\paragraph{Implementation Details.}
We utilize Qwen3-VL-4B-Instruct~\cite{Qwen3-VL} as our base model. For SFT, we perform full-parameter fine-tuning on the GDP-26K training set with a maximum sequence length of 4096. For RLVR, we use the ROLL framework~\cite{roll} with GRPO~\cite{deepseekmath} on a curated subset of 2,000 training samples, using a learning rate of $1\times10^{-6}$, group size 8, and global batch size 128.

\subsection{Main Results}
\label{subsec:main_results}

Table~\ref{tab:main_results} and Table~\ref{tab:solid_results} summarize the parsing performance on PGDP-2K and SGDP-1K, respectively. Overall, our GDP-4B models achieve the best performance across both benchmarks, demonstrating that the proposed geometry formal language and training pipeline substantially enhance geometric perception beyond general MLLMs.

\paragraph{Results on PGDP.}
On the plane geometry benchmark, our GDP-4B-RL achieves a SOTA score of 96.4, significantly surpassing large-scale MLLMs. While models like GPT and Gemini-3-Flash perform well on basic primitives (e.g., \textit{Points}), they exhibit noticeable performance drops on \textit{Lines} and \textit{Semantic relations}. For instance, despite its massive scale, Qwen3-VL-235B-Thinking achieves only 72.4 F1 on Semantics, whereas our model attains 90.7. This substantial gap underscores that general visual pre-training is insufficient for capturing explicit geometric logic, a capability effectively unlocked by our specialized formal training.

\paragraph{Results on SGDP.}
The challenge of geometry perception is more evident in solid geometry, where most baselines struggle significantly with \textit{Lines}, \textit{Circles}, and \textit{Planes} compared to their PGDP performance. Due to the strong requirement for spatial understanding, even strong models like GPT-5.2 achieve only 72.8 on \textit{Lines}, 65.3 on \textit{Circles} and 75.9 on \textit{Planes}. In contrast, GDP-4B-RL demonstrates robust spatial understanding, maintaining high precision across all primitives and achieving an overall score of 94.9. These results confirm that our framework successfully bridges the gap in solid geometry parsing, enabling the precise perception where general MLLMs fail.

\paragraph{Effect of RLVR.}
The comparison between GDP-4B-SFT and GDP-4B-RL highlights the critical role of verifiable reinforcement learning. We observe that for fundamental primitives such as points and lines, the performance gains from RLVR are relatively marginal, as the SFT model already achieves near-saturated accuracy in these basic perception tasks. In contrast, RLVR demonstrates its primary strength in refining higher-order structures: it boosts the semantics score by 3.2\% on PGDP and the plane F1-score by 1.5\% on SGDP. This suggests that the reward signal specifically incentivizes the model to transcend simple visual recognition, effectively resolving complex ambiguities and ensuring overall geometric consistency.

\begin{table*}[t]
    \centering
    \small 
    
    % 稍微压缩一点列间距以适应增加的列
    \setlength{\tabcolsep}{6.5pt}
    
    \begin{tabular}{l cccc c cc}
        \toprule
        \multirow{2}{*}{\textbf{Model}} & \multicolumn{4}{c}{\textbf{Plane Geometry}} & & \multicolumn{2}{c}{\textbf{Solid Geometry}} \\
        \cmidrule(lr){2-5} \cmidrule(lr){7-8}
         & \textbf{GeoQA} & \textbf{PGPS9K} & \textbf{Geometry3K} & \textbf{MathVerse} & & \textbf{SolidGeo} & \textbf{MathVerse} \\
        \midrule
        
        Ministral-3-8B & 39.6 & 41.2 & 44.8 & 51.2 & & 9.6 & 26.0 \\
        \quad \textbf{+ Ours} & \textbf{41.5}\textcolor{red}{~\scalebox{0.9}{(+1.9)}} & \textbf{47.3}\textcolor{red}{~\scalebox{0.9}{(+6.1)}} & \textbf{53.3}\textcolor{red}{~\scalebox{0.9}{(+8.5)}} & \textbf{52.4}\textcolor{red}{~\scalebox{0.9}{(+1.2)}} & & 8.8\textcolor{darkgreen}{~\scalebox{0.9}{(-0.6)}} & \textbf{26.8}\textcolor{red}{~\scalebox{0.9}{(+0.8)}} \\
        \midrule
        
        Qwen3-VL-8B & 48.9 & 44.9 & 50.1 & 66.8 & & 59.0 & 42.0 \\
        \quad \textbf{+ Ours} & 48.7\textcolor{darkgreen}{~\scalebox{0.9}{(-0.2)}} & \textbf{53.9}\textcolor{red}{~\scalebox{0.9}{(+9.0)}} & \textbf{60.2}\textcolor{red}{~\scalebox{0.9}{(+10.1)}} & \textbf{68.5}\textcolor{red}{~\scalebox{0.9}{(+1.7)}} & & \textbf{62.1}\textcolor{red}{~\scalebox{0.9}{(+3.1)}} & \textbf{44.5}\textcolor{red}{~\scalebox{0.9}{(+2.5)}} \\
        \midrule
        
        Qwen2.5-VL-32B & 59.7 & 38.1 & 46.3 & 54.9 & & 52.5 & 36.1 \\
        \quad \textbf{+ Ours} & \textbf{61.7}\textcolor{red}{~\scalebox{0.9}{(+2.0)}} & \textbf{46.8}\textcolor{red}{~\scalebox{0.9}{(+8.7)}} & \textbf{55.8}\textcolor{red}{~\scalebox{0.9}{(+9.5)}} & 54.9\textcolor{gray}{~\scalebox{0.9}{(+0.0)}} & & \textbf{53.8}\textcolor{red}{~\scalebox{0.9}{(+1.3)}} & 34.4\textcolor{darkgreen}{~\scalebox{0.9}{(-1.7)}} \\
        \midrule
        
        Qwen3-VL-32B & 67.8 & 69.4 & 73.0 & 73.8 & & 73.7 & 45.3 \\
        \quad \textbf{+ Ours} & \textbf{70.6}\textcolor{red}{~\scalebox{0.9}{(+2.8)}} & \textbf{78.0}\textcolor{red}{~\scalebox{0.9}{(+8.6)}} & \textbf{82.6}\textcolor{red}{~\scalebox{0.9}{(+9.6)}} & \textbf{75.9}\textcolor{red}{~\scalebox{0.9}{(+2.1)}} & & \textbf{73.9}\textcolor{red}{~\scalebox{0.9}{(+0.2)}} & \textbf{47.0}\textcolor{red}{~\scalebox{0.9}{(+1.7)}} \\
        \midrule
        
        GPT-5.2-1211 & 55.3 & 78.1 & 84.5 & 76.3 & & 60.5 & 64.7 \\
        \quad \textbf{+ Ours} & \textbf{58.8}\textcolor{red}{~\scalebox{0.9}{(+3.5)}} & \textbf{82.2}\textcolor{red}{~\scalebox{0.9}{(+4.1)}} & \textbf{86.4}\textcolor{red}{~\scalebox{0.9}{(+1.9)}} & \textbf{77.8}\textcolor{red}{~\scalebox{0.9}{(+1.5)}} & & \textbf{61.3}\textcolor{red}{~\scalebox{0.9}{(+0.8)}} & \textbf{66.3}\textcolor{red}{~\scalebox{0.9}{(+1.6)}} \\
        
        \bottomrule
    \end{tabular}
    \caption{Downstream reasoning accuracy (\%) on Plane and Solid geometry benchmarks. We compare vanilla MLLMs with those augmented by our formal parsing (\textbf{+ Ours}). $\Delta$ (in \textcolor{red}{red}) denotes the absolute improvement. MathVerse results are reported on its plane and solid geometry subsets, respectively.}
    \label{tab:main_reasoning_results}
\end{table*}

\subsection{Diagram-level Exact Match Evaluation}
While category-level F1 measures fine-grained parsing quality, it does not necessarily indicate that a diagram is parsed perfectly as a whole. To better evaluate holistic correctness, we additionally report Sample Accuracy \textbf{(SA)} for each category and Perfect Parsing Rate \textbf{(PPR)} for the full diagram.

As shown in Table~\ref{tab:sa_ppr_combined}, strong general-purpose MLLMs may achieve competitive F1 scores on individual categories, yet their exact-match performance remains much lower at the sample level. This gap reflects a clear multiplier effect: even a single error in any primitive or semantic relation can invalidate the entire formal description. In contrast, our GDP-RL framework substantially mitigates this issue, achieving much higher SA and PPR on both plane and solid geometry. In particular, GDP-4B-RL reaches a PPR of 72.8\% on PGDP-2K and 70.9\% on SGDP-1K, demonstrating that our method not only improves fine-grained parsing quality, but also produces holistically correct formal outputs much more reliably.

\subsection{Downstream Geometry Reasoning}
\label{subsec:reasoning}

Having established the superior accuracy of our parsing model, we investigate its practical utility by using the parsed formal descriptions for downstream geometry reasoning task. Table~\ref{tab:main_reasoning_results} reports the performance of various MLLMs augmented with our parsing results across both plane and solid geometry benchmarks.

As observed, augmenting MLLMs with our formal parsing yields consistent improvements, particularly in plane geometry. On visually complex benchmarks like Geometry3K and PGPS9K, Qwen3-VL-8B achieves substantial gains of +10.1\% and +9.0\%, respectively, and even the advanced GPT-5.2 sees a solid +4.1\% boost. We attribute this to the high visual semantic density of these diagrams, where explicit parsing captures subtle constraints (e.g., parallelism, angles) essential for reasoning. In solid geometry, incorporating parsed primitives yields moderate yet positive gains. This narrower margin likely stems from two primary factors: (i) \textbf{Textual Explicitness}, where current solid geometry benchmarks often feature problem statements that already explicitly describe the geometric structure, leaving less "new" information for the parser to provide; and (ii) \textbf{Intrinsic Semantic Sparsity}, as solid geometry diagrams tend to contain fewer implicit symbolic constraints compared to their planar counterparts.

\subsection{Impact of Representation Form}
\label{subsec:ablation}
\begin{figure}[t]
    \centering
    \includegraphics[width=1.0\linewidth]{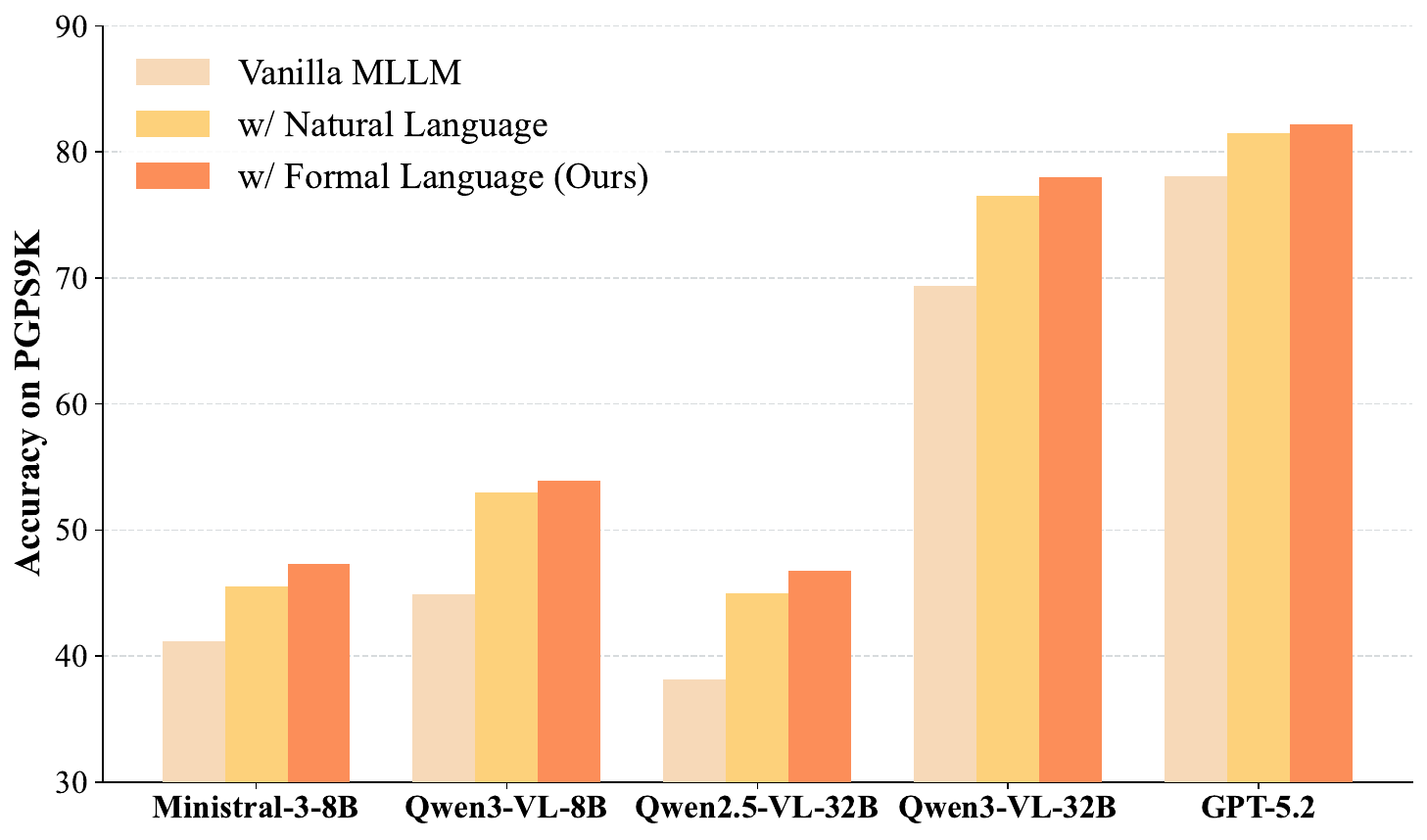}
    \caption{Effect of Representation Forms on PGPS9K Reasoning Accuracy.}
    \vspace{-3mm}
\label{fig:nlvsformal}
\end{figure}

To isolate the impact of the parsed geometry description format on geometry reasoning, we compare our Formal Language (FL) against Natural Language (NL) on the PGPS9K benchmark. To ensure strict semantic equivalence, we employ Gemini-3-Pro to translate our parsed formal sequences into coherent NL descriptions, ensuring the two forms differ only in representation. As illustrated in Figure~\ref{fig:nlvsformal}, while both augmentation strategies improve over the vanilla baseline, FL consistently outperforms NL in assisting geometric reasoning across all five evaluated models. This superiority suggests that compact, symbolic representations provide higher information density and a stronger inductive bias for geometric reasoning compared to verbose textual descriptions.

%% file: sec/7_conclusion.tex
\section{Conclusion}
\label{sec:conclusion}

In this work, we address the perception bottleneck in multimodal geometric reasoning by establishing a unified formal language and a parsing framework for both plane and solid geometry. We introduce the GDP-29K dataset, which effectively fills the critical data void in the solid geometry domain and significantly expands image diversity by incorporating both printed and hand-drawn styles. By employing a training paradigm that combines SFT with Reinforcement Learning via Verifiable Rewards, we ensure the syntactic rigor and geometric consistency of the generated formal descriptions. Experimental results demonstrate that our method achieves SOTA parsing performance, and the parsed formal descriptions serve as a vital cognitive scaffold, significantly boosting downstream geometry reasoning capabilities on benchmarks such as Geometry3K, PGPS9K, and MathVerse.

%% file: sec/limitations.tex
\section*{Limitations}
\label{sec:limitations}

While the GDP-29K dataset and our parsing framework establish a strong baseline, we acknowledge several limitations that guide future research. First, the current formal definitions within GDP-29K do not explicitly distinguish between visible and invisible (e.g., dashed) elements in solid geometry; incorporating explicit visibility attributes could further enhance the depth of solid geometry comprehension and spatial understanding. Second, the visual semantics of our current solid geometry samples are relatively sparse, primarily focusing on basic primitives. Future work aims to construct datasets with richer semantic diversity and more intricate spatial scenarios to further push the boundaries of fine-grained spatial understanding in multimodal models.

%% file: appendix/X_suppl.tex
\clearpage
\appendix

\section{More Details of GDP-29K}
\label{appendix:dataset}

In this section, we provide extended details regarding the GDP-29K dataset, including its manual collection process, formal language syntax, and comprehensive statistical analysis.

\subsection{Data Collection}
GDP-29K is specifically designed to address the lack of diversity and 3D coverage in existing geometry parsing benchmarks.

\paragraph{Handwritten Subset.} 
Our handwritten plane geometry subset is entirely manually drawn. We recruited 10 annotators with diverse handwriting styles to recreate 5,516 geometric diagrams using digital tablets and styluses. This process captures authentic stroke dynamics, varying line thicknesses, and realistic distortions (e.g., imperfect circles and non-straight lines). This high-fidelity data ensures that models trained on GDP-29K possess robust generalization capabilities for real-world educational scenarios, such as grading student sketches.

\paragraph{Solid Geometry Collection.} 
The solid geometry samples cover a wide spectrum of 3D structures, including prisms, pyramids, cones, cylinders, and frustums. These diagrams were curated from high-quality geometry textbooks and competitive math examinations. Each diagram was then meticulously annotated with our unified formal language to capture both its topological structure and spatial semantics.

\subsection{Detailed Statistical Analysis of GDP-29K}
We performed a comprehensive statistical analysis of the structures and semantic constraints within the GDP-29K dataset to verify its diversity and coverage.

\begin{figure}[htbp]
    \centering
    \includegraphics[width=1.0\linewidth]{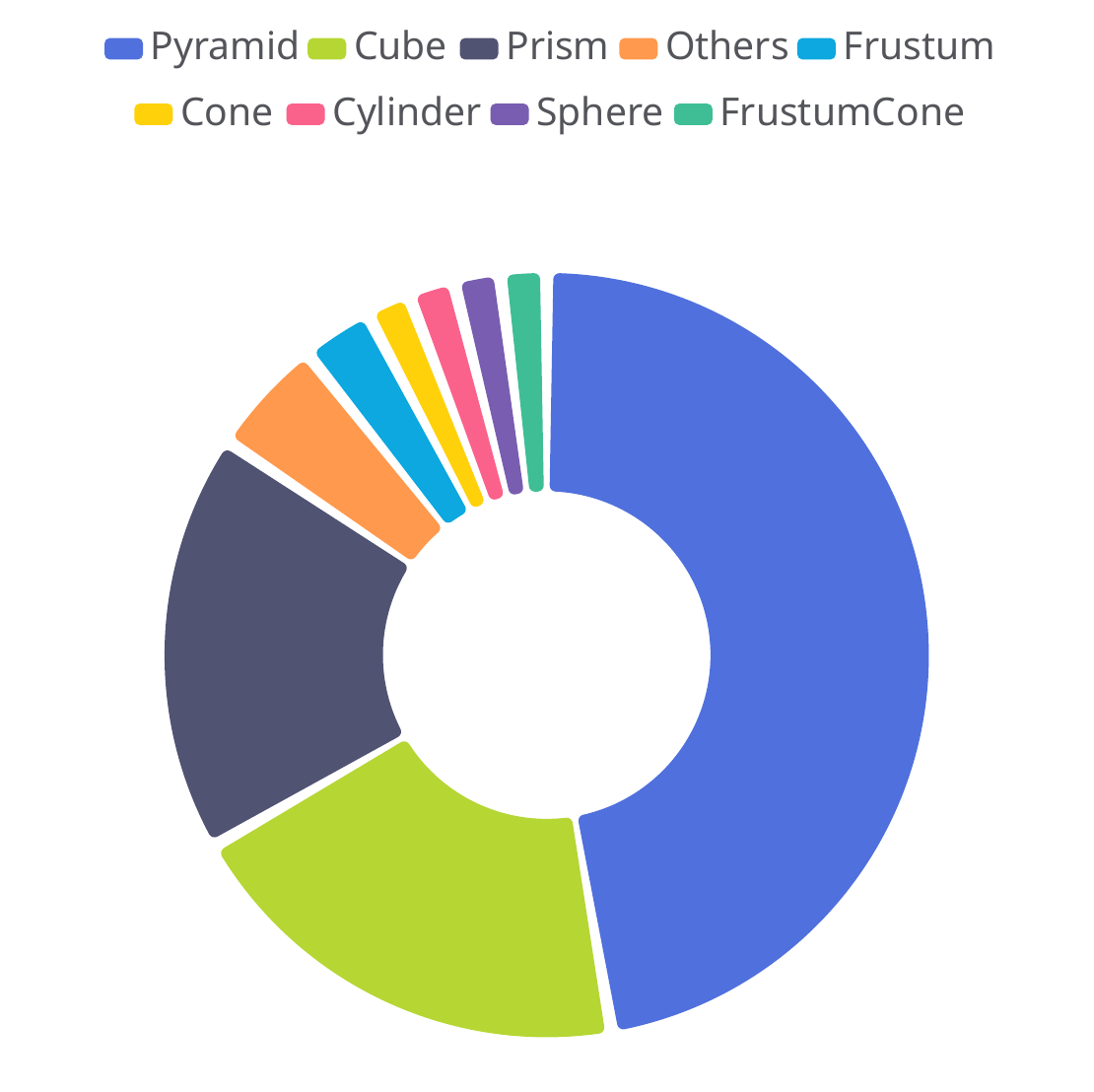}
    \caption{Distribution of 3D structures in the SGDP subset ($N=9,012$). The dataset covers a wide range of polyhedral forms and rotational solids to facilitate robust spatial perception.}
    \label{fig:sgdp_dist}
\end{figure}

\paragraph{Structural Diversity.} 
As illustrated in Figure~\ref{fig:sgdp_dist}, the SGDP subset (comprising 7,960 analyzed 3D samples) exhibits a rich variety of geometric structures. \textbf{Pyramids} constitute the largest portion of the dataset with 3,937 instances (49.46\%), reflecting their high frequency in 3D geometry problems. \textbf{Cubes} (1,618, 20.33\%) and \textbf{Prisms} (1,473, 18.51\%) follow as the next most prevalent categories. To ensure the model generalizes to complex and curved surfaces, the dataset incorporates \textbf{Frustums} (248, 3.12\%), \textbf{Cones} (156, 1.96\%), and \textbf{Cylinders} (82, 1.03\%), as well as rarer structures like \textbf{Spheres} (25, 0.31\%) and \textbf{Conic Frustums} (11, 0.14\%). A small percentage of \textbf{Others} (410, 5.15\%) includes hybrid or irregular solids. This distribution ensures that our model is exposed to both common polyhedral forms and more challenging rotational solids.

\begin{figure}[htbp]
    \centering
    \includegraphics[width=1.0\linewidth]{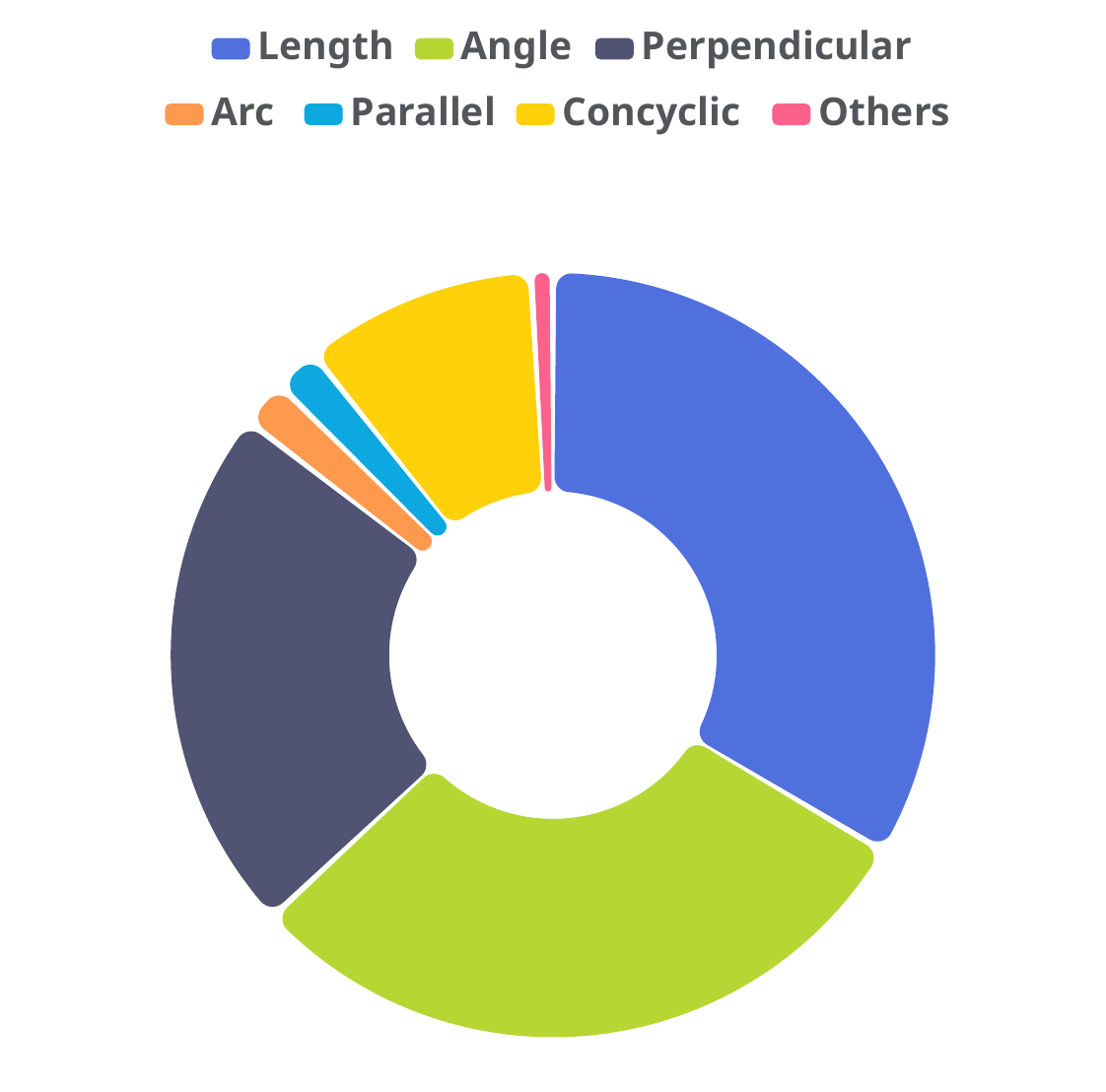}
    \caption{Distribution of semantic predicates across the PGDP subset ($N=48,613$). Metric constraints and orthogonal relations form the core of the geometric reasoning tasks.}
    \label{fig:semantics_dist}
\end{figure}

\paragraph{Semantic Richness.} 
The distribution of semantic constraints in the PGDP subset (Figure~\ref{fig:semantics_dist}) highlights the dataset's focus on rigorous logical relations. Out of 48,613 identified constraints, \textbf{Length} measurements (18,247, 37.54\%) and \textbf{Angle} specifications (16,067, 33.05\%) are the most prevalent, providing the metric foundation for geometric reasoning. Notably, \textbf{Perpendicularity} (\texttt{\textbackslash perp}) accounts for a significant 12,181 instances (25.06\%), emphasizing the importance of topological connectivity and orthogonal relations in theorem proving. Furthermore, \textbf{Arc} measures (1,074, 2.21\%) and \textbf{Parallelism} (1,037, 2.13\%) enrich the dataset by ensuring holistic coverage of plane geometry properties.

\section{Details of Data Annotation}
\label{appendix:annotation}

In alignment with the labeling strategy described in the main text, this section provides further specifics regarding our annotation workforce, the three-tier quality control protocol, and the redundancy filtering process.

\subsection{Annotation Workforce and Training}
We recruited 30 undergraduate students majoring in STEM fields (Science, Technology, Engineering, and Mathematics) to perform the annotation tasks. All participants underwent a standardized training session to familiarize themselves with our formal language's syntax and the 3D spatial relationship definitions. To ensure consistency, each annotator was required to pass a preliminary test consisting of 50 samples before contributing to the final dataset.

\subsection{Three-tier Quality Control Protocol}
To maintain a high standard of structural rigor, we implemented a rigorous three-tier workflow:

\begin{enumerate}
    \item \textbf{Annotation Stage:} 
    \begin{itemize}[leftmargin=*]
        \item \textit{Plane Geometry:} Annotators reviewed and corrected initial drafts provided by GPT-5. The primary focus was on fixing vertex ordering and ensuring all geometric constraints (e.g., parallelism) were captured.
        \item \textit{Solid Geometry:} Since MLLMs often fail to perceive 3D depth, annotators manually identified all faces, edges, and spatial relations from scratch, following the hierarchical structure of our formal language.
    \end{itemize}
    \item \textbf{Verification Stage:} A different student from the team acted as a peer reviewer for each annotated sample. They cross-checked the formal description against the original diagram to identify any missing primitives or incorrect semantic tags. Any discrepancies were returned to the original annotator for revision.
    \item \textbf{Final Acceptance Stage:} Our expert leads (authors of this study) performed a final audit on the verified samples. This stage focused on ensuring the logical consistency of the formal language and the accuracy of complex 3D structures (e.g., non-trivial frustums and spheroids). Only samples with 100\% consensus were moved to the final pool.
\end{enumerate}

\subsection{Redundancy Filtering and Final Statistics}
After the manual annotation, we performed a structural de-duplication step to enhance dataset diversity. We identified samples with \textbf{identical formal descriptions}—defined as instances where all primitives, semantic values, and topological relations were isomorphic—and retained only one representative image per structure.

Following this filtering process, the dataset was finalized at \textbf{28,977} high-quality samples. The distribution of these samples ensures that the model learns to generalize across diverse geometric layouts without being biased by repetitive structural patterns.

\section{Formal Language Specification}
As shown in Table~\ref{tab:syntax_details}, our formal language is characterized by its structural conciseness and a quasi-natural language style, intentionally designed to facilitate more effective understanding and generation by MLLMs. By adopting a syntax that mirrors both standard mathematical notation and intuitive linguistic phrasing (e.g., \texttt{AB \textbackslash perp to CD on X}), we reduce the mapping complexity from visual features to symbolic representations. This alignment leverages the model's pre-trained linguistic knowledge, ensuring that the formalization is not only mathematically rigorous but also highly accessible for model learning and reasoning.

\begin{table*}[h]
\centering
\small
\begin{tabular}{lp{6cm}p{7.5cm}}
\toprule
\textbf{Category} & \textbf{Formal Syntax (Example)} & \textbf{Geometric Description} \\ \midrule
\multirow{4}{*}{\textbf{Primitives}} & \texttt{point A} & Defines a point vertex named A. \\
 & \texttt{line A B C} & A line segment passing through points A, B, and C. \\
  & \texttt{line k lineson A B C} & A line lieson point A B C \\
 & \texttt{plane A B C D} & A plane defined by vertices A, B, C, and D. \\
 & \texttt{\textbackslash odot O lieson A B C} & A circle with center O passing through points A, B, and C. \\ \midrule
\multirow{3}{*}{\textbf{Semantics}} & \texttt{AB = 57} & The length of segment $AB$ is $57$. \\
 & \texttt{m \textbackslash angle ABC = 41} & The measure of $\angle ABC$ is $41^\circ$. \\
 & \texttt{m \textbackslash widehat AB = 90} & The angular measure of arc $AB$ is $90^\circ$. \\ \midrule
\multirow{2}{*}{\textbf{Relations}} & \texttt{AB \textbackslash perp to CD on X} & Line $AB$ is perpendicular to $CD$, intersecting at point $X$. \\
 & \texttt{AB \textbackslash parallel CD} & Line $AB$ is parallel to line $CD$. \\ \midrule
\multirow{8}{*}{\textbf{3D Solids}} & \texttt{solid Cube ABCD-A\_\{1\}B\_\{1\}C\_\{1\}D\_\{1\}} & A cube defined by its bottom and top faces. \\
 & \texttt{solid Prism ABC-A\_\{1\}B\_\{1\}C\_\{1\}} & A prism defined by its bottom and top faces. \\
 & \texttt{solid Frustum ABC-A\_\{1\}B\_\{1\}C\_\{1\}} & A frustum defined by its bottom and top bases. \\
 & \texttt{solid Pyramid O-ABC} & A pyramid defined by apex $O$ and base $ABC$. \\
 & \texttt{solid Spheriod O-ABCD} & A spheroid defined by center $O$ and surface points $A, B, C, D$. \\
 & \texttt{solid Cylinder AD-BC} & A cylinder defined by two lateral side segments $AD$ and $BC$. \\
 & \texttt{solid Cone P-OA} & A cone defined by apex $P$, base center $O$, and base point $A$. \\
 & \texttt{solid FrustumCone AD-BC} & A conical frustum defined by its lateral side segments. \\ \bottomrule
\end{tabular}
\caption{Detailed syntax and examples of the formal language in GDP-29K, covering 2D primitives and 3D solid structures.}
\label{tab:syntax_details}
\end{table*}

\paragraph{Key Design Principles.}
\begin{itemize}[leftmargin=*]
    \item \textbf{Topological Precision:} Beyond simple detection, our language explicitly denotes intersection points (e.g., \texttt{on X} in perpendicular relations). This provides the model with clear topological anchors, which is crucial for building a consistent geometric graph.
    \item \textbf{Semantic Intuition:} By adopting a quasi-natural language style (e.g., using keywords like \texttt{lieson}, \texttt{perp to}, and \texttt{parallel}), we align the formal syntax with the model's pre-trained linguistic priors. This reduces the cognitive load on the MLLM during the translation from pixels to symbols.
    \item \textbf{Hierarchical Composition:} 3D solids are not treated as isolated entities but are composed of 2D primitives (points, lines, and planes). This design ensures a unified representational space, allowing the model to leverage its 2D parsing experience when tackling complex 3D structures.
\end{itemize}

\section{Hierarchical Prompting Strategy}
\label{subsec:prompt_strategy}

To accurately bridge the gap between raw geometric images and rigorous formal symbolic language, we propose a hierarchical prompting strategy. We decompose the formalization task into five specialized, decoupled modules: \texttt{point\_line}, \texttt{circle}, \texttt{plane}, \texttt{solid}, and \texttt{semantic}. The detailed design of these prompts is illustrated in Figure~\ref{fig:prompt_points_and_lines}, \ref{fig:prompt_circles}, \ref{fig:prompt_semantics}, \ref{fig:prompt_solid_structure}, and \ref{fig:prompt_planes}.

\paragraph{Structural Layer: Primitive Extraction.}
The first four prompts focus on extracting the "topological skeleton" of the diagram. 
(1) \textbf{\texttt{point\_line}}: This template identifies all labeled points and their collinearity, enforcing strict ordering to maintain the physical continuity of lines. 
(2) \textbf{\texttt{circle}}: It guides the model to locate centers and discrete points on circumferences, ensuring a clear distinction between the boundary and the interior. 
(3) \textbf{\texttt{plane}} and (4) \textbf{\texttt{solid}}: These prompts provide spatial context, where the former handles 2D regional layouts and the latter focuses on 3D volumetric structures, such as identifying hidden edges and face-to-face connectivity in polyhedra.

\paragraph{Logical Layer: Semantic Constraint Mapping.}
(5) \textbf{\texttt{semantic}}: Building upon the structural skeleton, this template extracts logical relationships. It instructs the model to parse explicit visual markers (e.g., right-angle squares, parallel arrows) into formal clauses (e.g., $\perp$, $\parallel$). By isolating semantic reasoning from primitive detection, we prevent the model from making unfounded visual assumptions and ensure that every generated clause is grounded in explicit symbolic evidence.

\paragraph{Capability Elicitation and Fair Comparison.}
The core rationale behind this hierarchical decomposition is to maximize the potential of various MLLMs. Geometric formalization is a high-cognitive-load task; by adopting a "divide-and-conquer" approach, we alleviate the instruction-following burden on the models, allowing them to focus on granular sub-tasks. Furthermore, this standardized prompting framework ensures a fair comparison across different model architectures. It eliminates the confounding factor of models' varying abilities to handle multi-step formatting in a single pass, instead providing a uniform interface to evaluate their true underlying geometric perception capabilities.

\begin{figure}[htbp]
    \centering
    \begin{promptbox}{Prompt:  Points \& Lines}
    You are an expert in geometry diagram structure analysis. \\
    Your SOLE task is to identify the Points and Lines from the image.

    1. \textbf{Points}: \\
    \quad - Identify all labeled points in the diagram. \\
    \quad - Format: $[A, B, C, A_{1}\dots]$

    2. \textbf{Lines}: \\
    \quad - Identify all lines (Including solid lines and dashed lines). \\
    \quad - A "line" must include \textbf{ALL} labeled points lying on that straight segment. \\
    \quad - Do NOT split a line into smaller segments. If $A, B, C$ are collinear, output ONE line \texttt{line A B C}, NOT \texttt{line A B} and \texttt{line B C}. \\
    \quad - Do NOT skip intermediate points. If $B$ is between $A$ and $C$, you MUST write \texttt{line A B C}, NOT \texttt{line A C}. \\
    \quad - Points must be listed in the strict order they appear visually (from one end to the other). \\
    \quad - Format: \texttt{line P1 P2 P3 P4 P5}

    \vspace{0.5em}
    \textbf{\#\#\# Output Format Template} \\
    \textbf{Points:} \\
    $[list\_ of\_points]$ \\
    \textbf{Lines:} \\
    \texttt{line P1 P2 P3 P4 P5} \\
    \texttt{line P6 P7}
    \end{promptbox}
    
    \caption{Prompts for geometric structural analysis. These prompts guide the model to extract the topological skeleton of the diagram.}
    \label{fig:prompt_points_and_lines}
\end{figure}

\begin{figure}[t]
    \centering
    \begin{promptbox}{Prompt: Circles}
    You are an expert in geometry diagram structure analysis. \\
    Your SOLE task is to identify \textbf{Circles} from the image.

    \vspace{0.5em}
    \textbf{\#\#\# Definitions and Rules}

    1. \textbf{Circles}: \\
    - Identify all circles (or major arcs acting as circles) in the diagram. \\
    - \textbf{Structure}: For each circle, you must identify: \\
    a. The \textbf{Center} point. \\
    b. \textbf{Points on Circumference}: List \textbf{ALL} labeled points that lie strictly \textbf{ON} the curve. \\
    - \textbf{Critical Constraints}: \\
    a. Do not miss any point that on the circle's boundary. \\
    b. Do NOT include points that are \textit{inside} or \textit{outside} the circle (except the Center). Only list points on the rim. \\
    - \textbf{Format}: \texttt{\textbackslash odot Center lieson P1 P2 P3}

    2. \textbf{No Circles}: \\
    - If there are no circles in the diagram, leave the section under ``\textbf{Circles:}'' empty or write ``None''.

    \vspace{0.5em}
    \textbf{\#\#\# Output Format Template} \\
    \textbf{Circles:} \\
    \texttt{\textbackslash odot O lieson A B C M $A_{1}$}
    \end{promptbox}
    \caption{Prompts for geometric structural analysis. These prompts guide the model to extract the topological skeleton of the diagram.}
    \label{fig:prompt_circles}
\end{figure}

\begin{figure}[htbp]
    \centering
    \begin{promptbox}{Prompt: Semantics}
    You are an expert in geometry semantic analysis. \\
    Your task is to extract geometric relationships, equations, and constraints from the image text and symbols.

    \vspace{0.5em}
    \textbf{\#\#\# Semantic Clauses Templates} \\
    You must use the following templates.

    1. \textbf{Perpendicular}: \\
    \quad - Template: \texttt{AB \textbackslash perp CD on P} \\
    \quad - Note: You MUST include ``on P''.

    2. \textbf{Parallel}: \\
    \quad - Template: \texttt{AB \textbackslash parallel CD}

    3. \textbf{Angle Measure \& Equations}: \\
    \quad - Template: \texttt{m \textbackslash angle ABC = 30} or \texttt{m \textbackslash angle ABC = 2x + 5}

    4. \textbf{Segment Lengths \& Congruence}: \\
    \quad - Template: \texttt{AB = 5} or \texttt{AB = CD}

    5. \textbf{Arc Measure}: \\
    \quad - Template: \texttt{m \textbackslash widehat AB = 60}

    \vspace{0.5em}
    \textbf{\#\#\# Constraints \& Anti-Redundancy Rules}

    1. \textbf{Collinear Points Rule}: \\
    \quad - If A-B-C are collinear and perpendicular to D-E, output ONE representative clause (e.g., \texttt{AC \textbackslash perp DE on B}). \\
    \quad - Do NOT list \texttt{AB \textbackslash perp DE}, \texttt{BC \textbackslash perp DE} separately.

    2. \textbf{Right Angles}: \\
    \quad - If a square symbol is present, use \texttt{AB \textbackslash perp CD on B}. \\
    \quad - Do NOT output \texttt{m \textbackslash angle ABC = 90} if you output the perpendicular clause.

    3. \textbf{Strict Symbolism (No Visual Assumptions)}: \\
    \quad - \textbf{Do NOT} infer relationships based on visual appearance (e.g., lines that ``look'' parallel or perpendicular). \\
    \quad - \textbf{Perpendicularity}: ONLY output \texttt{\textbackslash perp} if there is an explicit right-angle symbol (square marker) or text declaration. \\
    \quad - \textbf{Parallelism}: ONLY output \texttt{\textbackslash parallel} if there are explicit arrow markers on the lines or text declaration. \\
    \quad - Only output relationships that are \textbf{explicitly displayed} in the diagram.

    \vspace{0.5em}
    \textbf{Output Format Template} \\
    \textbf{Semantic Clauses:} \\
    \textit{Clause 1} \\
    \textit{Clause 2} \\
    \dots
    \end{promptbox}
    
    \caption{Prompts for geometric structural analysis. These prompts guide the model to extract the topological skeleton of the diagram.}
    \label{fig:prompt_semantics}
\end{figure}

\begin{figure}[htbp]
    \centering
    \begin{promptbox}{Prompt: Solid Structure}
    You are an expert in solid geometry diagram parsing. \\
    Your SOLE task is to classify the \textbf{3D Structure} of the geometric figure.

    \vspace{0.5em}
    \textbf{\#\#\# Allowed Categories (Strictly Choose One)} \\
    You must classify the solid into exactly ONE of the following categories:

    \quad 1. \textbf{Pyramid} \quad (General pyramids, e.g., triangular/rectangular) \\
    \quad 2. \textbf{Prism} \quad (General prisms, e.g., triangular/hexagonal) \\
    \quad 3. \textbf{Cube} \quad (Regular hexahedron, all faces are squares) \\
    \quad 4. \textbf{Frustum} \quad (Truncated pyramid) \\
    \quad 5. \textbf{Cylinder} \quad (Circular cylinder) \\
    \quad 6. \textbf{Cone} \quad (Circular cone) \\
    \quad 7. \textbf{FrustumCone} \quad (Truncated cone) \\
    \quad 8. \textbf{Spheroid} \quad (Ball shape)

    \vspace{0.5em}
    \textbf{\#\#\# Naming Rules} \\
    \quad - After the category keyword, you MUST append the labeled vertices. \\
    \quad - \textbf{Format}: \texttt{Category Vertices} \\
    \quad - \textit{Examples}: \\
    \quad \quad \texttt{Pyramid P-ABCD} \\
    \quad \quad \texttt{Cube ABCD-$A_{1}B_{1}C_{1}D_{1}$} \\
    \quad \quad \texttt{Cylinder $O_{1}$-$O_{2}$} \\
    \quad \quad \texttt{Spheroid O}

    \vspace{0.5em}
    \textbf{Output Format Template} \\
    \textbf{Structure:} \\
    \texttt{Pyramid P-ABC}
    \end{promptbox}
    \caption{Prompts for 3D geometric structural analysis. These prompts guide the model to classify the solid type and identify its defining vertices.}
    \label{fig:prompt_solid_structure}
\end{figure}

\begin{figure}[htbp]
    \centering
    \begin{promptbox}{Prompt: Planes}
    You are an expert in solid geometry diagram parsing. \\
    Your ONLY task: output ALL \textbf{planes} visible or implied in the structure. \\
    Do NOT output any explanation.

    \vspace{0.5em}
    \textbf{\#\#\# What counts as a ``Plane''} \\
    \quad 1. \textbf{Boundary Faces}: The external flat surfaces (top, bottom, sides). \\
    \quad 2. \textbf{Internal Sections}: Explicitly drawn planes cutting through the solid.

    \vspace{0.5em}
    \textbf{\#\#\# Rules \& Constraints (CRITICAL)} \\
    \quad 1. \textbf{Maximal Point Set Principle}: \\
    \quad \quad - List \textbf{EVERY} labeled point on the plane (vertices, edge points, etc.). \\
    \quad \quad - \textbf{Do NOT} output a plane if it is a subset of another output. \\
    \quad \quad - \textit{Example}: Use \texttt{plane A B E C} instead of \texttt{plane A B C} if E is on BC. \\
    \quad 2. \textbf{Handle Hidden Faces}: \\
    \quad \quad - Infer hidden faces based on structure and dashed lines. \\
    \quad \quad - Only list faces that have at least 3 labeled points. \\
    \quad 3. \textbf{Output Format}: \\
    \quad \quad - Begin with header \texttt{**Planes:**} on the first line. \\
    \quad \quad - Format: \texttt{plane P1 P2 P3 ...}

    \vspace{0.5em}
    \textbf{Output Format Template} \\
    \textbf{Planes:} \\
    \texttt{plane A B C D} \\
    \texttt{plane A B E} \\
    \texttt{plane C D F G}
    \end{promptbox}
    \caption{Prompts for solid geometric plane extraction. These prompts enforce the Maximal Point Set Principle to ensure all coplanar labeled points are grouped together.}
    \label{fig:prompt_planes}
\end{figure}

% \begin{figure}[htbp]
%     \centering

%     \begin{promptbox}{Prompt: Circles}
%     \textbf{Circles}: Identify centers and all points on circumference.\\
%     \quad - \textbf{Strict Rule}: Do NOT miss points on boundary; exclude interior/exterior points.\\
%     \quad - \textit{Format}: \texttt{\textbackslash odot Center lieson P1 P2 P3}
%     \end{promptbox}
%     \caption{Prompts for geometric structural analysis. These prompts guide the model to extract the topological skeleton of the diagram.}
%     \label{fig:prompt_structure}
% \end{figure}

\section{Data Examples}
\label{sec:data_examples}

\begin{figure*}[htbp]
    \centering
    \includegraphics[width=1.0\linewidth]{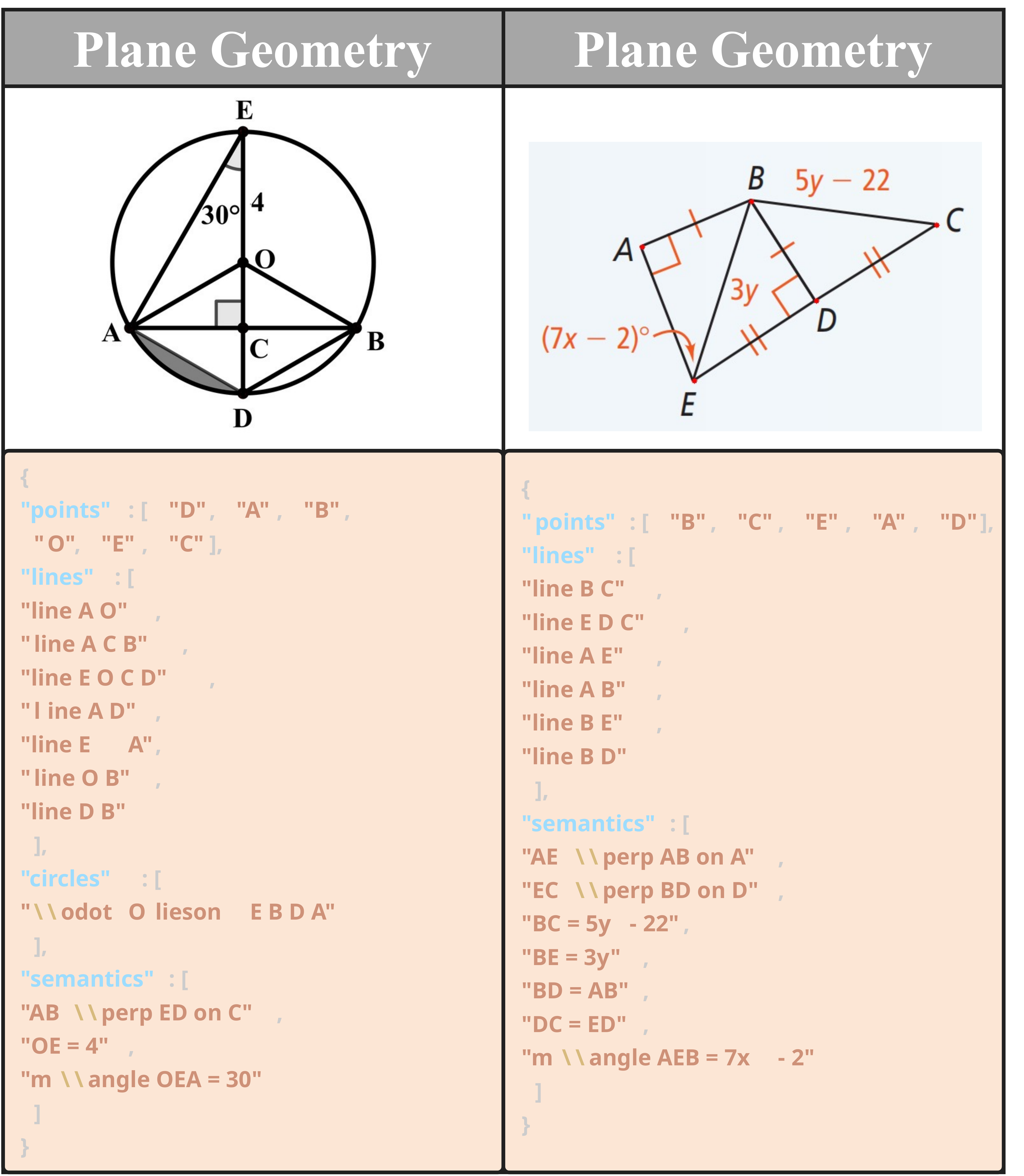}
    \caption{Representative plane geometry samples from the GDP-29K dataset.}
    \label{fig:plane1}
\end{figure*}

\begin{figure*}[htbp]
    \centering
    \includegraphics[width=1.0\linewidth]{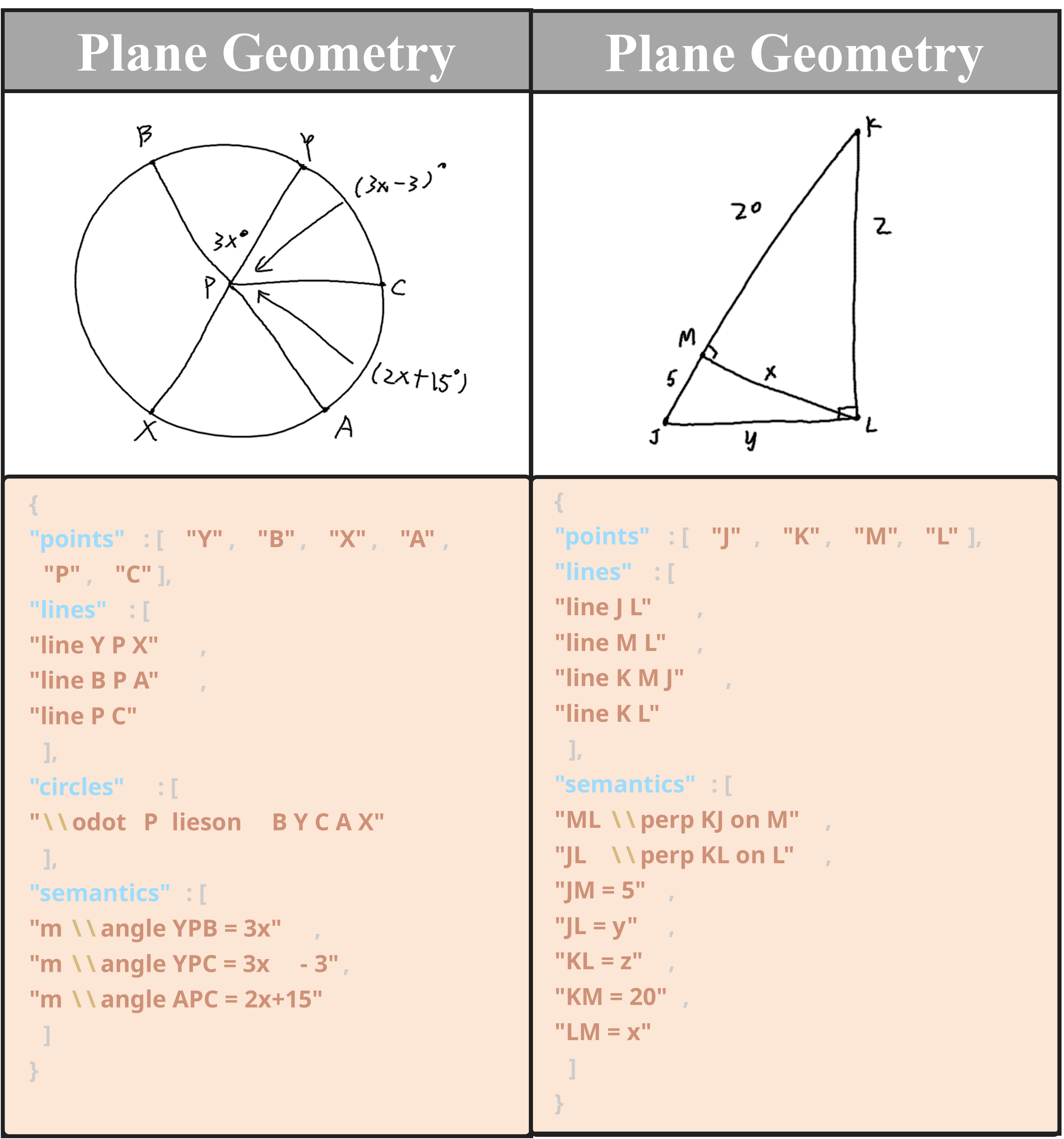}
    \caption{Representative plane geometry samples from the GDP-29K dataset.}
    \label{fig:plane2}
\end{figure*}

\begin{figure*}[htbp]
    \centering
    \includegraphics[width=1.0\linewidth]{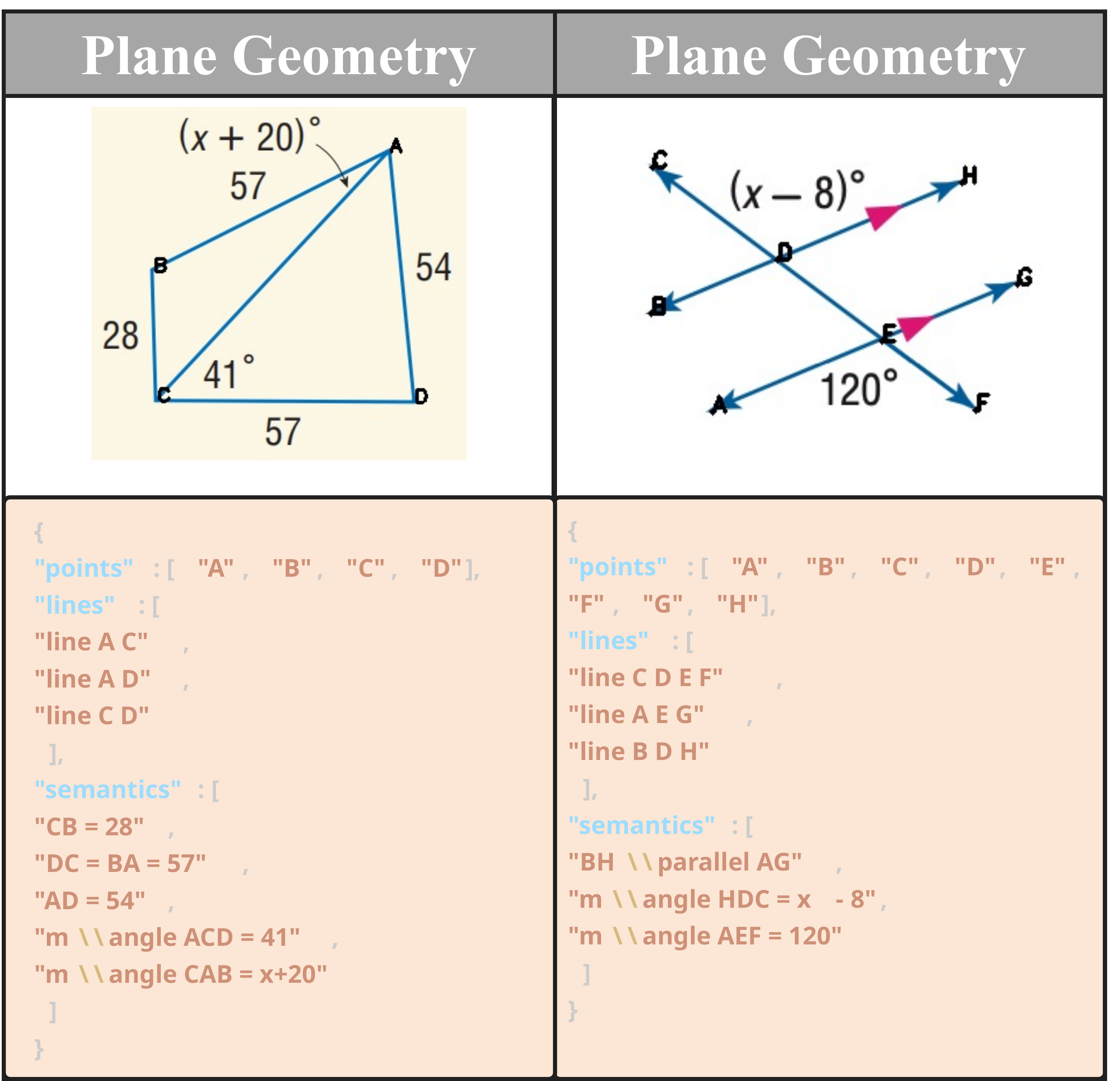}
    \caption{Representative plane geometry samples from the GDP-29K dataset.}
    \label{fig:plane3}
\end{figure*}

To provide a concrete illustration of the GDP-29K dataset, we present representative examples from both the planar and solid geometry subsets in Figures~\ref{fig:plane1}, \ref{fig:plane2}, and \ref{fig:plane3}. These examples demonstrate the capability of our unified formal language to bridge the gap between visual diagrams and symbolic logic.

As shown in the examples, for plane geometry, our formalization accurately captures fundamental primitives such as points, lines, and circles, while simultaneously encoding complex semantic constraints like perpendicularity markers and angle measures. For solid geometry, the parsed outputs successfully represent 3D structural skeletons, including the identification of hidden edges and the connectivity between polyhedral planes and vertices. Notably, our unified formal language is designed to be highly concise and follows a style that closely resembles natural language. This human-readable syntax ensures that the symbolic descriptions remain intuitive and interpretable, while effectively eliciting the logical reasoning capabilities of large multimodal models.

\section{Case Studies}
\label{appendix:case_studies}

\begin{figure*}[htbp]
    \centering
    \includegraphics[width=1.0\linewidth]{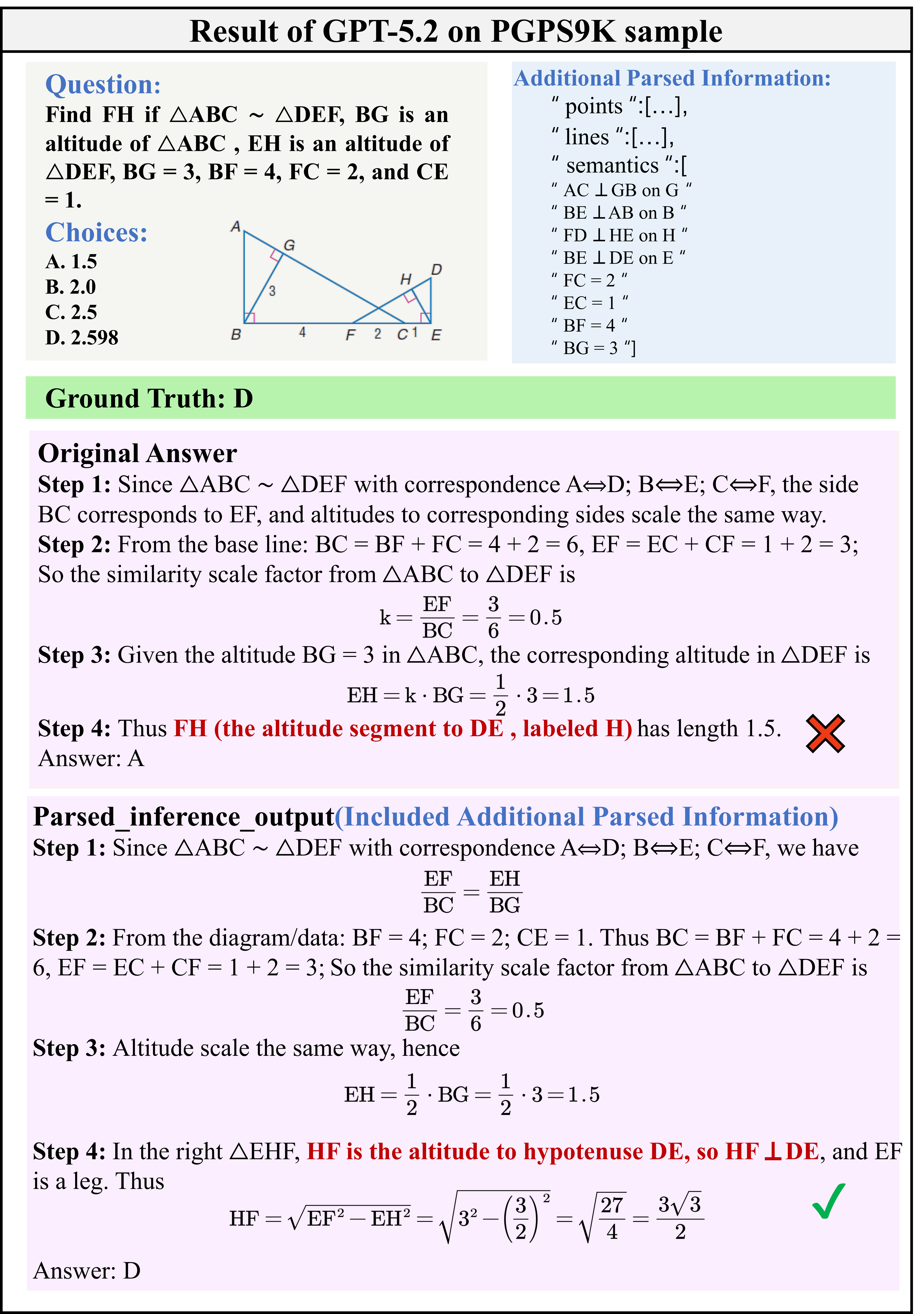}
    \caption{Qualitative comparison between Direct Inference and our method (+ Ours) on PGPS9K. Our formal parsing provides precise symbolic grounding that rectifies reasoning errors.}
    \label{fig:case1}
\end{figure*}

\begin{figure*}[htbp]
    \centering
    \includegraphics[width=1.0\linewidth]{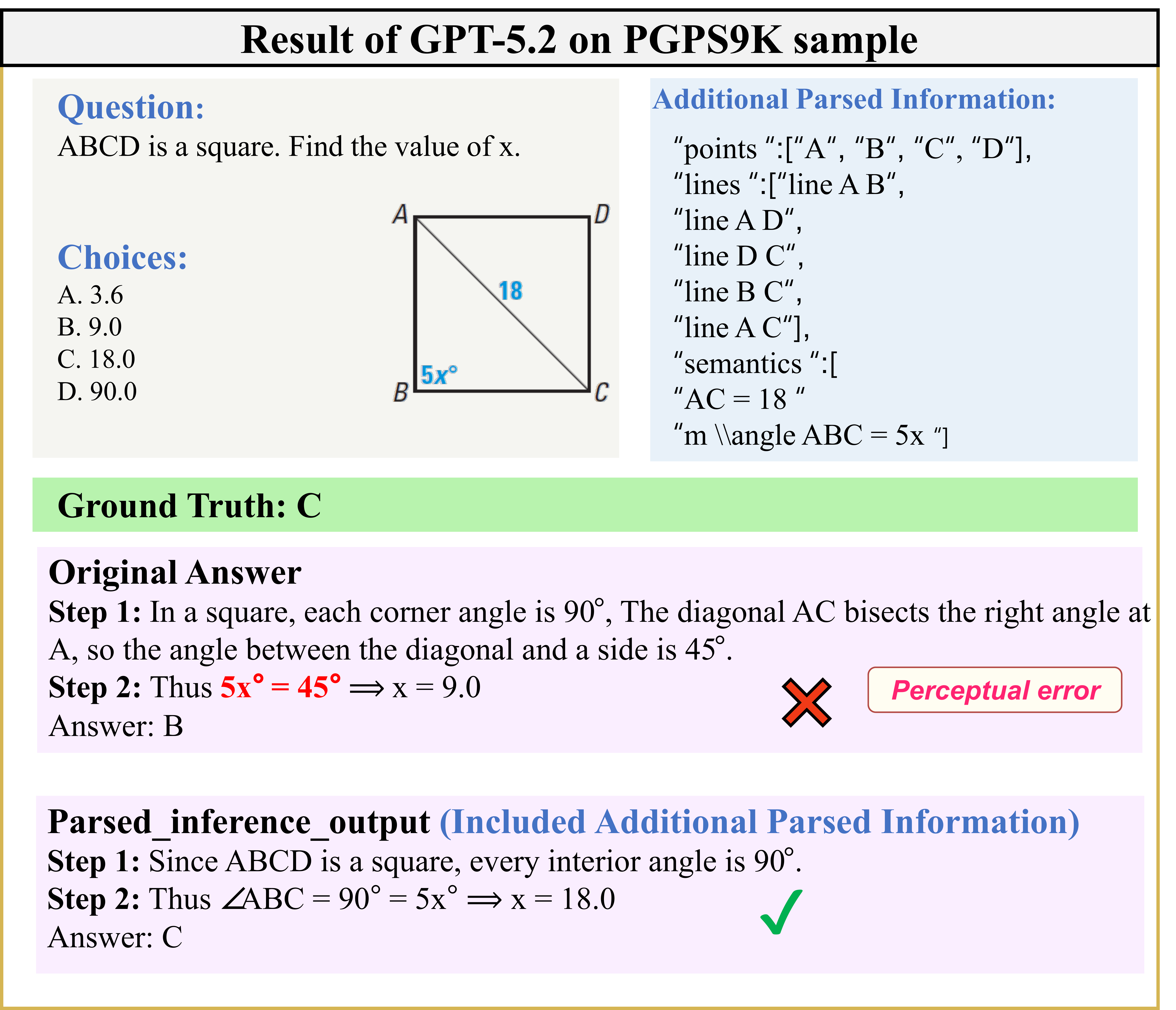}
    \caption{Qualitative comparison between Direct Inference and our method (+ Ours) on PGPS9K. Our formal parsing provides precise symbolic grounding that rectifies reasoning errors.}
    \label{fig:case2}
\end{figure*}

\begin{figure*}[htbp]
    \centering
    \includegraphics[width=1.0\linewidth]{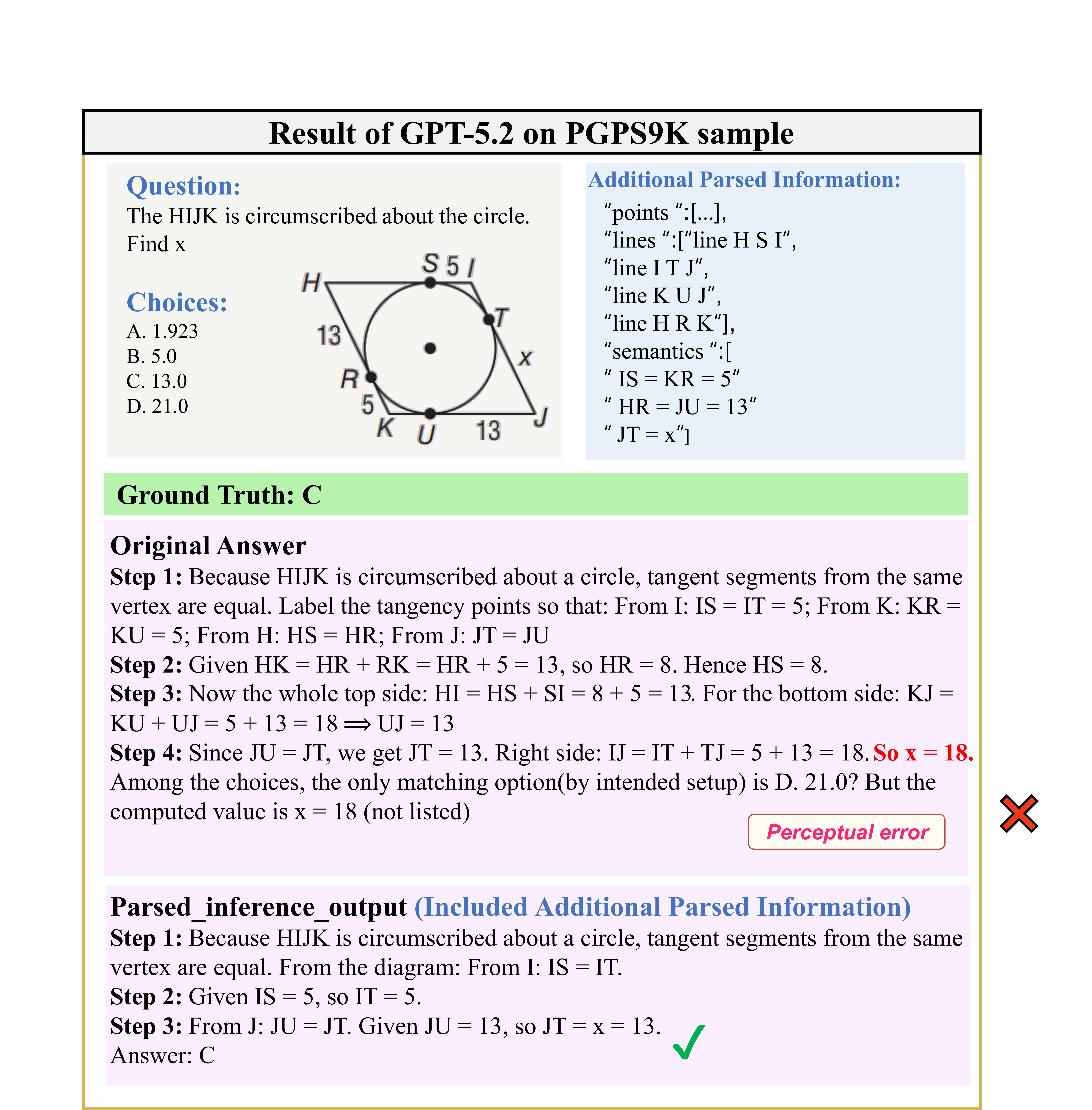}
    \caption{Qualitative comparison between Direct Inference and our method (+ Ours) on PGPS9K. Our formal parsing provides precise symbolic grounding that rectifies reasoning errors.}
    \label{fig:case3}
\end{figure*}

To illustrate the effectiveness of the formalized descriptions generated by our parsing method, we provide three qualitative examples in Figure~\ref{fig:case1},~\ref{fig:case2}, and~\ref{fig:case3}. These cases compare the performance of GPT-5.2-1211 on PGPS9K under two settings: \textbf{Direct Inference} and \textbf{+ Ours} (reasoning augmented by our GDP-4B formal parsing). 

As shown in the examples, our parsing results provide a precise symbolic foundation that corrects the model's reasoning trajectory, leading to the accurate final answer.

\begin{figure*}[htbp]
    \centering
    \includegraphics[width=1.0\linewidth]{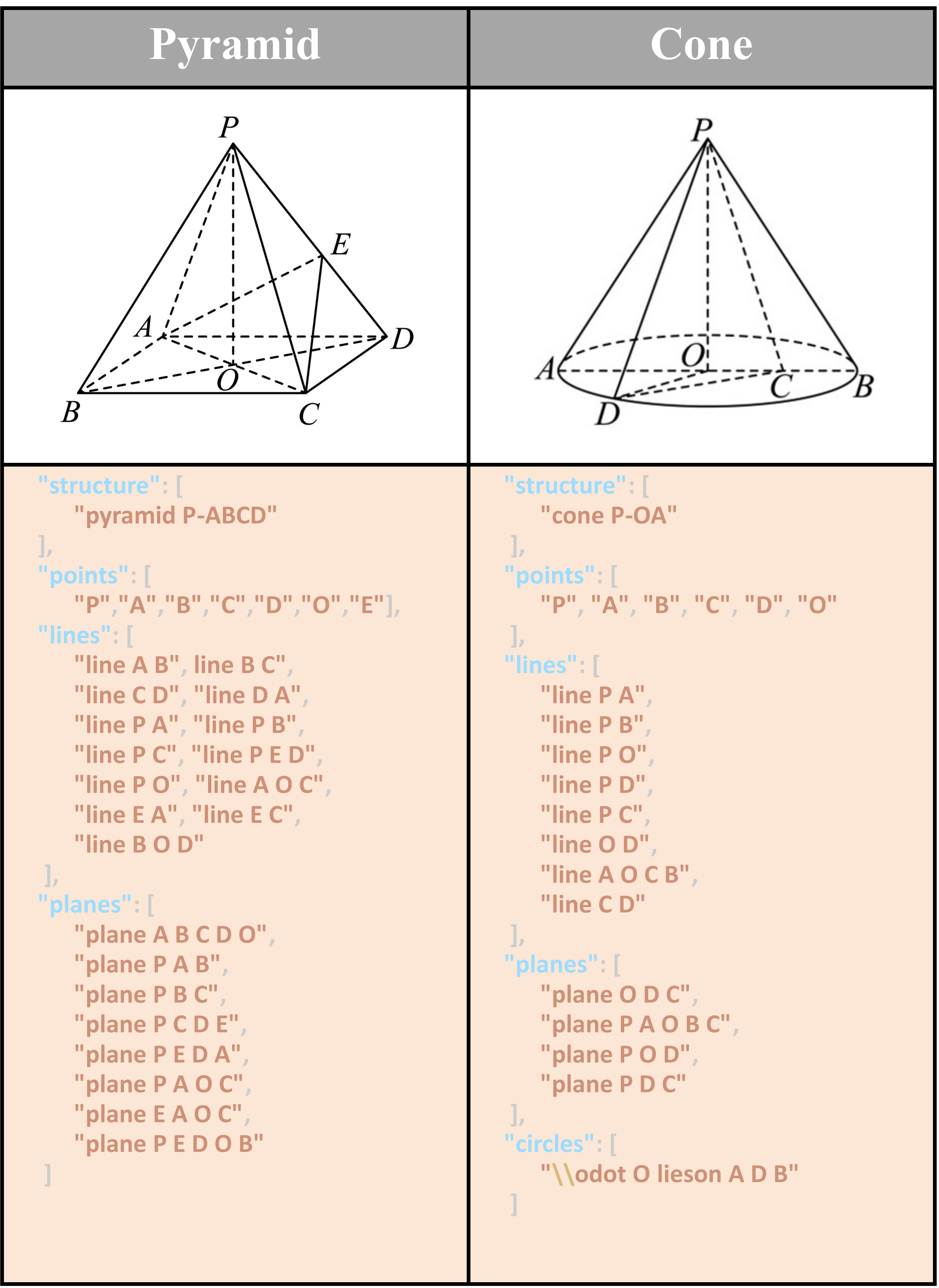}
    \caption{Representative solid geometry samples from the GDP-29K dataset.}
    \label{fig:solid1}
\end{figure*}

\begin{figure*}[htbp]
    \centering
    \includegraphics[width=1.0\linewidth]{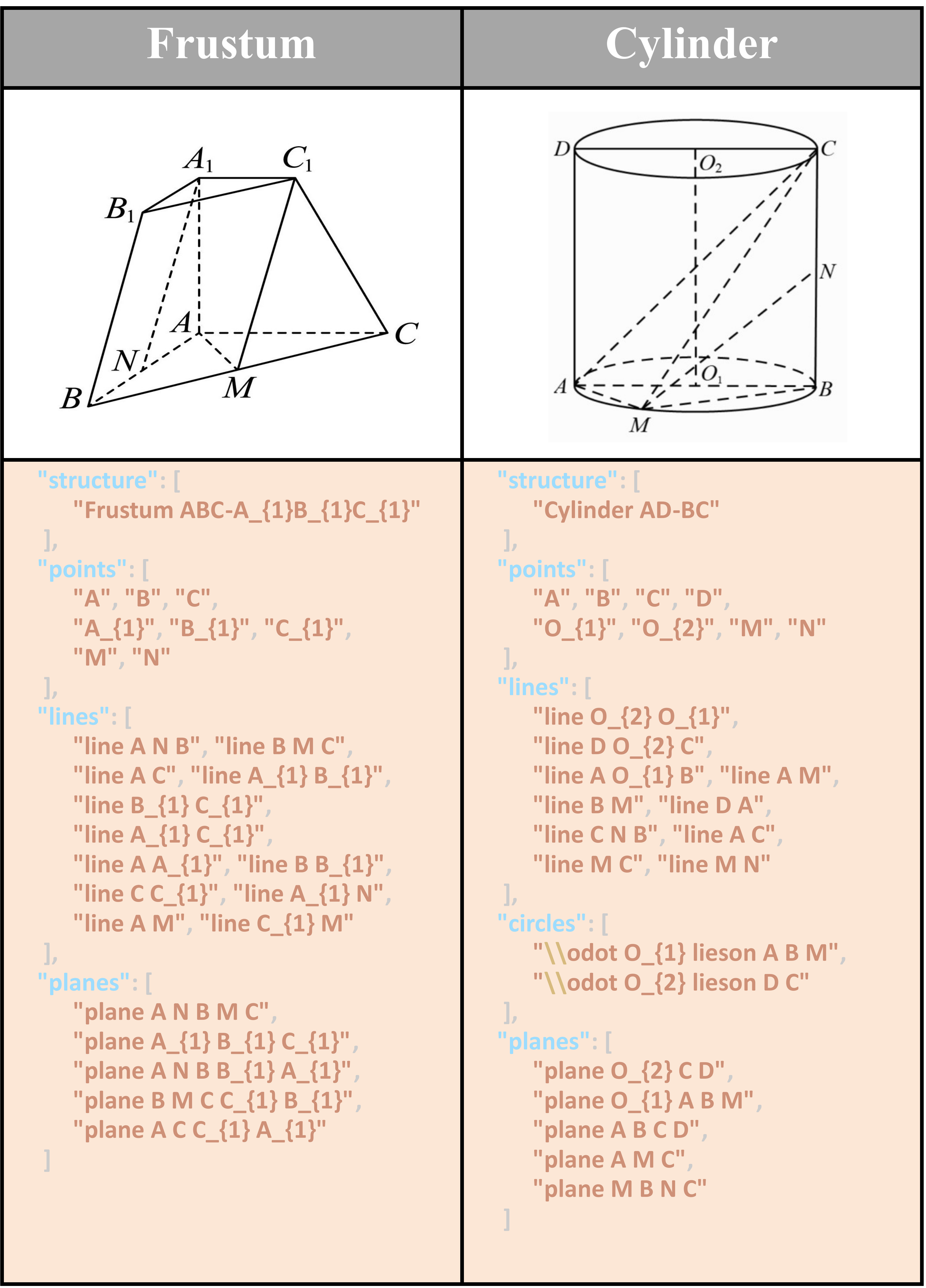}
    \caption{Representative solid geometry samples from the GDP-29K dataset.}
    \label{fig:solid2}
\end{figure*}

\begin{figure*}[htbp]
    \centering
    \includegraphics[width=1.0\linewidth]{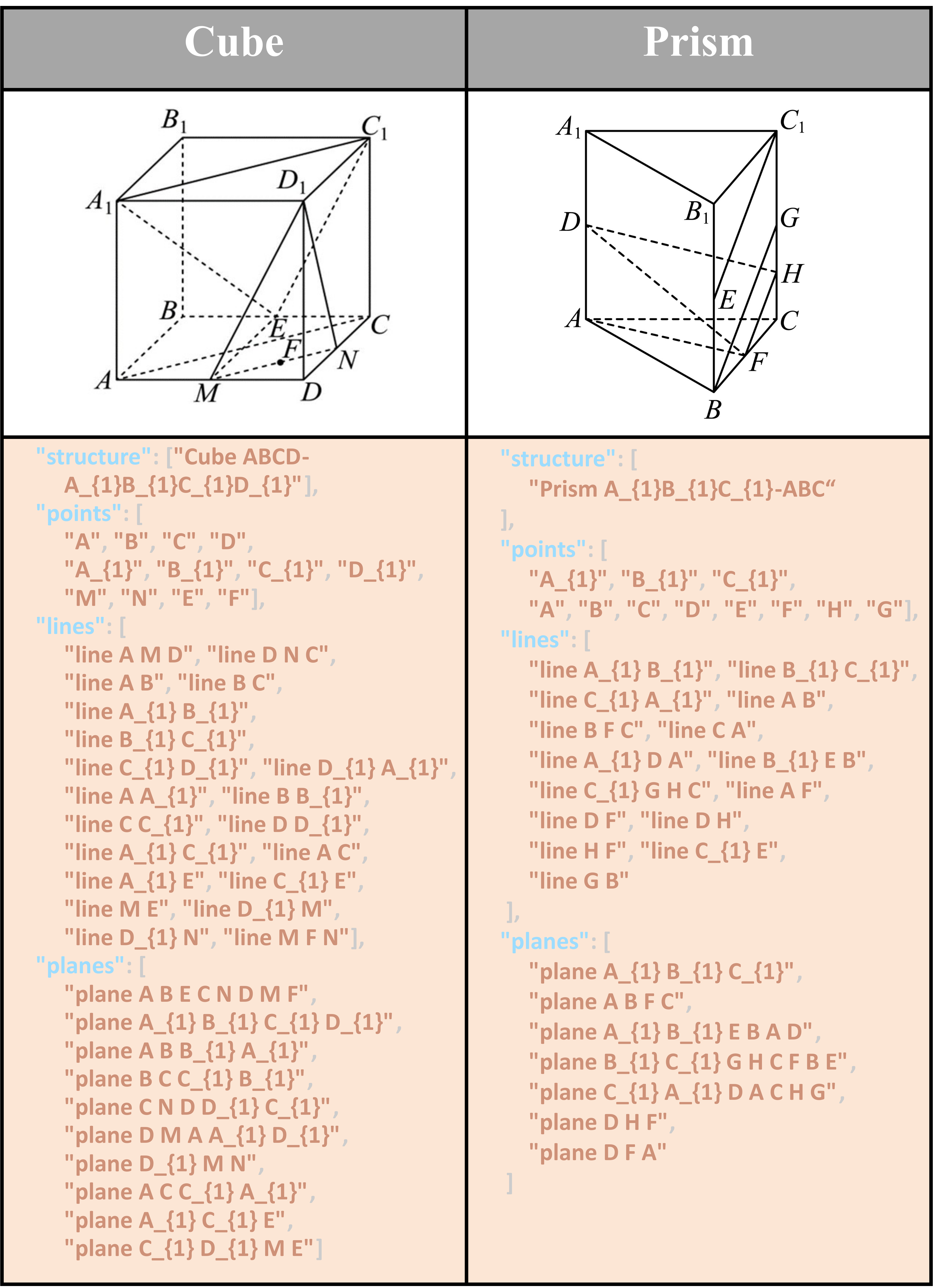}
    \caption{Representative solid geometry samples from the GDP-29K dataset.}
    \label{fig:solid3}
\end{figure*}

\begin{figure*}[htbp]
    \centering
    \includegraphics[width=1.0\linewidth]{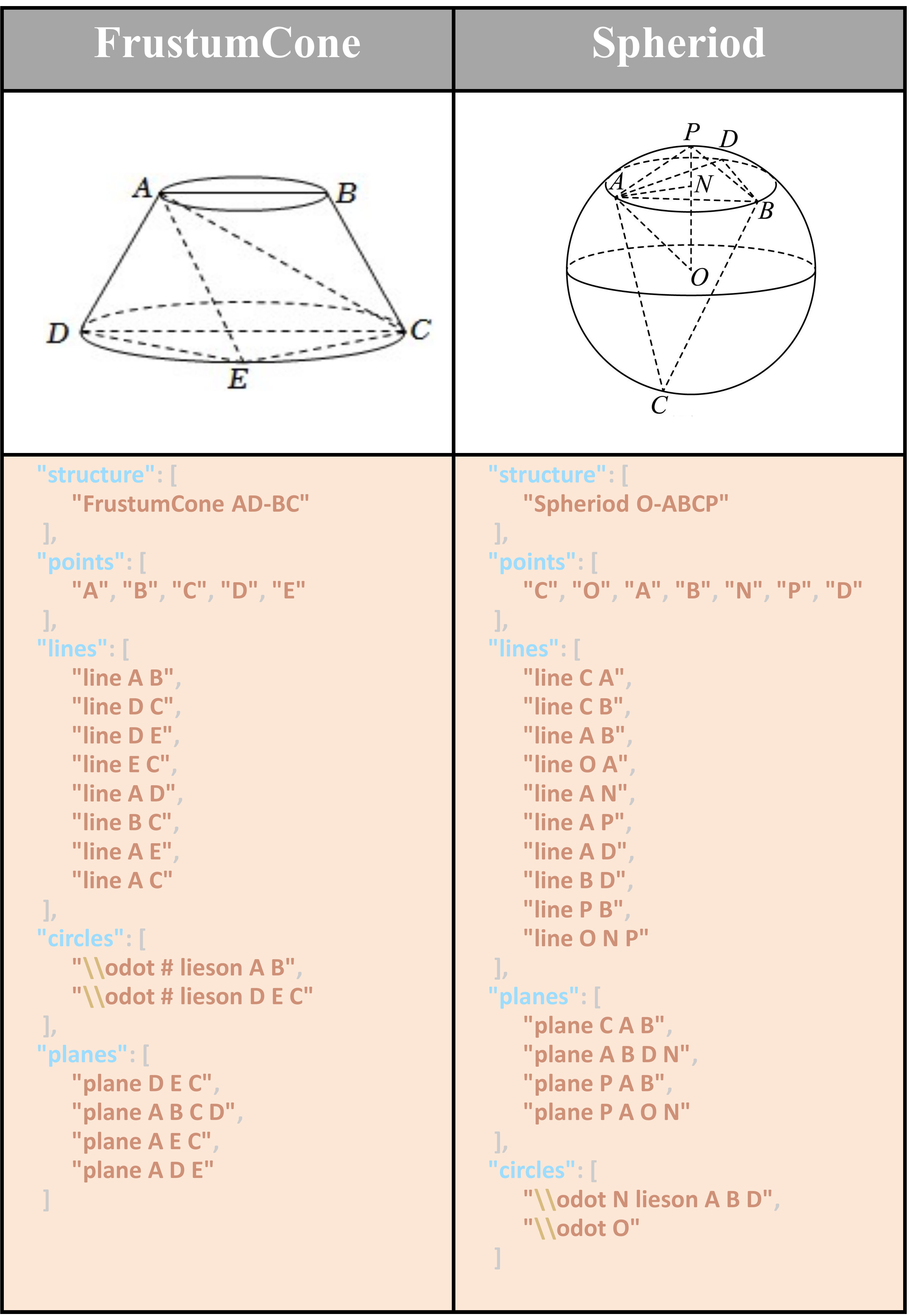}
    \caption{Representative solid geometry samples from the GDP-29K dataset.}
    \label{fig:solid4}
\end{figure*}

%% file: custom.bib
@book{riemannian,
  title={Riemannian geometry},
  author={Petersen, Peter},
  volume={171},
  year={2006},
  publisher={Springer}
}

@article{geoclass,
  title={On the relationship between plane and solid geometry},
  author={Arana, Andrew and Mancosu, Paolo},
  journal={The Review of Symbolic Logic},
  volume={5},
  number={2},
  pages={294--353},
  year={2012},
  publisher={Cambridge University Press}
}

@article{alphageometry,
  title={Solving olympiad geometry without human demonstrations},
  author={Trinh, Trieu H and Wu, Yuhuai and Le, Quoc V and He, He and Luong, Thang},
  journal={Nature},
  volume={625},
  number={7995},
  pages={476--482},
  year={2024},
  publisher={Nature Publishing Group UK London}
}

@article{chou1996automated,
  title={Automated generation of readable proofs with geometric invariants: I. Multiple and shortest proof generation},
  author={Chou, Shang-Ching and Gao, Xiao-Shan and Zhang, Jing-Zhong},
  journal={Journal of Automated Reasoning},
  volume={17},
  number={3},
  pages={325--347},
  year={1996},
  publisher={Springer}
}

@article{mathvision,
  title={Measuring multimodal mathematical reasoning with math-vision dataset},
  author={Wang, Ke and Pan, Junting and Shi, Weikang and Lu, Zimu and Ren, Houxing and Zhou, Aojun and Zhan, Mingjie and Li, Hongsheng},
  journal={Advances in Neural Information Processing Systems},
  volume={37},
  pages={95095--95169},
  year={2024}
}

@inproceedings{mathverse,
  title={MATHVERSE: Does Your Multi-modal LLM Truly See the Diagrams in Visual Math Problems?},
  author={Zhang, Renrui and Jiang, Dongzhi and Zhang, Yichi and Lin, Haokun and Guo, Ziyu and Qiu, Pengshuo and Zhou, Aojun and Lu, Pan and Chang, Kai-Wei and Qiao, Yu and others},
  booktitle={European Conference on Computer Vision},
  pages={169--186},
  year={2024}
}

@inproceedings{geoeval,
  title={GeoEval: Benchmark for Evaluating LLMs and Multi-Modal Models on Geometry Problem-Solving},
  author={Zhang, Jiaxin and Li, Zhong-Zhi and Zhang, Ming-Liang and Yin, Fei and Liu, Cheng-Lin and Moshfeghi, Yashar},
  booktitle={Findings of the Association for Computational Linguistics ACL 2024},
  pages={1258--1276},
  year={2024}
}

@article{solidgeo,
  title={SOLIDGEO: Measuring Multimodal Spatial Math Reasoning in Solid Geometry},
  author={Wang, Peijie and Yang, Chao and Li, Zhong-Zhi and Yin, Fei and Ran, Dekang and Tian, Mi and Ji, Zhilong and Bai, Jinfeng and Liu, Cheng-Lin},
  journal={arXiv preprint arXiv:2505.21177},
  year={2025}
}

@article{llavaonevision,
  title={LLaVA-OneVision: Easy Visual Task Transfer},
  author={Li, Bo and Zhang, Yuanhan and Guo, Dong and Zhang, Renrui and Li, Feng and Zhang, Hao and Zhang, Kaichen and Zhang, Peiyuan and Li, Yanwei and Liu, Ziwei and others},
  journal={Transactions on Machine Learning Research}
}

@article{internvl3.5,
  title={Internvl3. 5: Advancing open-source multimodal models in versatility, reasoning, and efficiency},
  author={Wang, Weiyun and Gao, Zhangwei and Gu, Lixin and Pu, Hengjun and Cui, Long and Wei, Xingguang and Liu, Zhaoyang and Jing, Linglin and Ye, Shenglong and Shao, Jie and others},
  journal={arXiv preprint arXiv:2508.18265},
  year={2025}
}

@article{Qwen3-VL,
      title={Qwen3-VL Technical Report}, 
      author={Shuai Bai and Yuxuan Cai and Ruizhe Chen and Keqin Chen and Xionghui Chen and Zesen Cheng and Lianghao Deng and Wei Ding and Chang Gao and Chunjiang Ge and Wenbin Ge and Zhifang Guo and Qidong Huang and Jie Huang and Fei Huang and Binyuan Hui and Shutong Jiang and Zhaohai Li and Mingsheng Li and Mei Li and Kaixin Li and Zicheng Lin and Junyang Lin and Xuejing Liu and Jiawei Liu and Chenglong Liu and Yang Liu and Dayiheng Liu and Shixuan Liu and Dunjie Lu and Ruilin Luo and Chenxu Lv and Rui Men and Lingchen Meng and Xuancheng Ren and Xingzhang Ren and Sibo Song and Yuchong Sun and Jun Tang and Jianhong Tu and Jianqiang Wan and Peng Wang and Pengfei Wang and Qiuyue Wang and Yuxuan Wang and Tianbao Xie and Yiheng Xu and Haiyang Xu and Jin Xu and Zhibo Yang and Mingkun Yang and Jianxin Yang and An Yang and Bowen Yu and Fei Zhang and Hang Zhang and Xi Zhang and Bo Zheng and Humen Zhong and Jingren Zhou and Fan Zhou and Jing Zhou and Yuanzhi Zhu and Ke Zhu},
	  journal={arXiv preprint arXiv:2511.21631},
      year={2025}
}

@article{survey1,
  title={A Survey of Deep Learning for Geometry Problem Solving},
  author={Ma, Jianzhe and Wang, Wenxuan and Jin, Qin},
  journal={arXiv preprint arXiv:2507.11936},
  year={2025}
}

@article{survey2,
  title={Towards Geometry Problem Solving in the Large Model Era: A Survey},
  author={Zhao, Yurui and Wang, Xiang and Liu, Jiahong and King, Irwin and Huang, Zhitao},
  journal={arXiv preprint arXiv:2506.02690},
  year={2025}
}

@misc{gemini3pro,
  author       = {{Google}},
  title        = {Gemini 3 Pro},
  howpublished = {\url{https://deepmind.google/models/gemini/pro/}},
  year         = {2025},
  month        = nov
}

@techreport{gpt5,
  author      = {OpenAI},
  title       = {{GPT-5 System Card}},
  institution = {OpenAI},
  year        = {2025},
  month       = aug,
  note        = {Version published August 7 2025. Available at: \url{https://openai.com/index/gpt-5-system-card/}},
  howpublished= {\url{https://openai.com/index/gpt-5-system-card/}}
}

@inproceedings{seo2014diagram,
  title={Diagram understanding in geometry questions},
  author={Seo, Min Joon and Hajishirzi, Hannaneh and Farhadi, Ali and Etzioni, Oren},
  booktitle={Proceedings of the Twenty-Eighth AAAI Conference on Artificial Intelligence},
  pages={2831--2838},
  year={2014}
}

@inproceedings{intergps,
  title={Inter-GPS: Interpretable Geometry Problem Solving with Formal Language and Symbolic Reasoning},
  author={Lu, Pan and Gong, Ran and Jiang, Shibiao and Qiu, Liang and Huang, Siyuan and Liang, Xiaodan and Zhu, Song-chun},
  booktitle={Proceedings of the 59th Annual Meeting of the Association for Computational Linguistics and the 11th International Joint Conference on Natural Language Processing (Volume 1: Long Papers)},
  pages={6774--6786},
  year={2021}
}

@inproceedings{pgdpnet,
  title={Plane Geometry Diagram Parsing},
  author={Zhang, Ming-Liang and Yin, Fei and Hao, Yi-Han and Liu, Cheng-Lin},
  booktitle={Proceedings of the Thirty-First International Joint Conference on Artificial Intelligence},
  pages={1636--1643},
  year={2022}
}

@article{formalgeo1,
  title={Formalgeo: An extensible formalized framework for olympiad geometric problem solving},
  author={Zhang, Xiaokai and Zhu, Na and He, Yiming and Zou, Jia and Huang, Qike and Jin, Xiaoxiao and Guo, Yanjun and Mao, Chenyang and Li, Yang and Zhu, Zhe and others},
  journal={arXiv preprint arXiv:2310.18021},
  year={2023}
}

@inproceedings{formalgeo2,
  title={Formal representation and solution of plane geometric problems},
  author={Zhang, Xiaokai and Zhu, Na and Qin, Cheng and Li, Yang and Zeng, Zhenbing and Leng, Tuo},
  booktitle={The 4th Workshop on Mathematical Reasoning and AI at NeurIPS'24},
  year={2024}
}

@inproceedings{pgps9k,
  title={A multi-modal neural geometric solver with textual clauses parsed from diagram},
  author={Zhang, Ming-Liang and Yin, Fei and Liu, Cheng-Lin},
  booktitle={Proceedings of the Thirty-Second International Joint Conference on Artificial Intelligence},
  pages={3374--3382},
  year={2023}
}

@inproceedings{mathvista,
  title={MathVista: Evaluating Mathematical Reasoning of Foundation Models in Visual Contexts},
  author={Lu, Pan and Bansal, Hritik and Xia, Tony and Liu, Jiacheng and Li, Chunyuan and Hajishirzi, Hannaneh and Cheng, Hao and Chang, Kai-Wei and Galley, Michel and Gao, Jianfeng},
  booktitle={The Twelfth International Conference on Learning Representations}
}

@inproceedings{wemath,
  title={We-math: Does your large multimodal model achieve human-like mathematical reasoning?},
  author={Qiao, Runqi and Tan, Qiuna and Dong, Guanting and MinhuiWu, MinhuiWu and Sun, Chong and Song, Xiaoshuai and Wang, Jiapeng and GongQue, Zhuoma and Lei, Shanglin and Zhang, Yifan and others},
  booktitle={Proceedings of the 63rd Annual Meeting of the Association for Computational Linguistics (Volume 1: Long Papers)},
  pages={20023--20070},
  year={2025}
}

@article{geosense,
  title={Geosense: Evaluating identification and application of geometric principles in multimodal reasoning},
  author={Xu, Liangyu and Zhao, Yingxiu and Wang, Jingyun and Wang, Yingyao and Pi, Bu and Wang, Chen and Zhang, Mingliang and Gu, Jihao and Li, Xiang and Zhu, Xiaoyong and others},
  journal={arXiv preprint arXiv:2504.12597},
  year={2025}
}

@inproceedings{mvmath,
  title={Mv-math: Evaluating multimodal math reasoning in multi-visual contexts},
  author={Wang, Peijie and Li, Zhong-Zhi and Yin, Fei and Ran, Dekang and Liu, Cheng-Lin},
  booktitle={Proceedings of the Computer Vision and Pattern Recognition Conference},
  pages={19541--19551},
  year={2025}
}

@inproceedings{geoqa,
  title={Geoqa: A geometric question answering benchmark towards multimodal numerical reasoning},
  author={Chen, Jiaqi and Tang, Jianheng and Qin, Jinghui and Liang, Xiaodan and Liu, Lingbo and Xing, Eric and Lin, Liang},
  booktitle={Findings of the Association for Computational Linguistics: ACL-IJCNLP 2021},
  pages={513--523},
  year={2021}
}

@inproceedings{gllava,
  title={G-LLaVA: Solving Geometric Problem with Multi-Modal Large Language Model},
  author={Gao, Jiahui and Pi, Renjie and Zhang, Jipeng and Ye, Jiacheng and Zhong, Wanjun and Wang, Yufei and HONG, Lanqing and Han, Jianhua and Xu, Hang and Li, Zhenguo and others},
  booktitle={The Thirteenth International Conference on Learning Representations}
}

@article{autogeo,
  title={AutoGeo: Automating Geometric Image Dataset Creation for Enhanced Geometry Understanding},
  author={Huang, Zihan and Wu, Tao and Lin, Wang and Zhang, Shengyu and Chen, Jingyuan and Wu, Fei},
  journal={IEEE Transactions on Multimedia},
  volume={27},
  pages={3105--3116},
  year={2025}
}

@inproceedings{mavis,
  title={MAVIS: Mathematical Visual Instruction Tuning with an Automatic Data Engine},
  author={Zhang, Renrui and Wei, Xinyu and Jiang, Dongzhi and Guo, Ziyu and Zhang, Yichi and Tong, Chengzhuo and Liu, Jiaming and Zhou, Aojun and Zhang, Shanghang and Gao, Peng and others},
  booktitle={The Thirteenth International Conference on Learning Representations}
}

@article{codeplot,
  title={Codeplot-cot: Mathematical visual reasoning by thinking with code-driven images},
  author={Duan, Chengqi and Sun, Kaiyue and Fang, Rongyao and Zhang, Manyuan and Feng, Yan and Luo, Ying and Liu, Yufang and Wang, Ke and Pei, Peng and Cai, Xunliang and others},
  journal={arXiv preprint arXiv:2510.11718},
  year={2025}
}

@inproceedings{geox,
  title={GeoX: Geometric Problem Solving Through Unified Formalized Vision-Language Pre-training},
  author={Xia, Renqiu and Li, Mingsheng and Ye, Hancheng and Wu, Wenjie and Zhou, Hongbin and Yuan, Jiakang and Peng, Tianshuo and Cai, Xinyu and Yan, Xiangchao and Wang, Bin and others},
  booktitle={The Thirteenth International Conference on Learning Representations}
}

@article{deepseekmath,
  title={Deepseekmath: Pushing the limits of mathematical reasoning in open language models},
  author={Shao, Zhihong and Wang, Peiyi and Zhu, Qihao and Xu, Runxin and Song, Junxiao and Bi, Xiao and Zhang, Haowei and Zhang, Mingchuan and Li, YK and others},
  journal={arXiv preprint arXiv:2402.03300},
  year={2024}
}

@article{deepseekr1,
  title={Deepseek-r1 incentivizes reasoning in llms through reinforcement learning},
  author={Guo, Daya and Yang, Dejian and Zhang, Haowei and Song, Junxiao and Wang, Peiyi and Zhu, Qihao and Xu, Runxin and Zhang, Ruoyu and Ma, Shirong and Bi, Xiao and others},
  journal={Nature},
  volume={645},
  number={8081},
  pages={633--638},
  year={2025},
  publisher={Nature Publishing Group UK London}
}

@article{eagle,
  title={Eagle: Elevating geometric reasoning through llm-empowered visual instruction tuning},
  author={Li, Zhihao and Du, Yao and Liu, Yang and Zhang, Yan and Liu, Yufang and Zhang, Mengdi and Cai, Xunliang},
  journal={arXiv preprint arXiv:2408.11397},
  year={2024}
}

@article{auxsolidmath,
  title={Geovlmath: Enhancing geometry reasoning in vision-language models via cross-modal reward for auxiliary line creation},
  author={Guo, Shasha and Pang, Liang and Wang, Xi and Wang, Yanling and Shen, Huawei and Zhang, Jing},
  journal={arXiv preprint arXiv:2510.11020},
  year={2025}
}

@inproceedings{olympiadbench,
  title={Olympiadbench: A challenging benchmark for promoting agi with olympiad-level bilingual multimodal scientific problems},
  author={He, Chaoqun and Luo, Renjie and Bai, Yuzhuo and Hu, Shengding and Thai, Zhen and Shen, Junhao and Hu, Jinyi and Han, Xu and Huang, Yujie and Zhang, Yuxiang and others},
  booktitle={Proceedings of the 62nd Annual Meeting of the Association for Computational Linguistics (Volume 1: Long Papers)},
  pages={3828--3850},
  year={2024}
}

@article{roll,
  title={Reinforcement Learning Optimization for Large-Scale Learning: An Efficient and User-Friendly Scaling Library},
  author={Wang, Weixun and Xiong, Shaopan and Chen, Gengru and Gao, Wei and Guo, Sheng and He, Yancheng and Huang, Ju and Liu, Jiaheng and Li, Zhendong and Li, Xiaoyang and others},
  journal={arXiv preprint arXiv:2506.06122},
  year={2025}
}

@inproceedings{sun2024determlr,
  title={Determlr: Augmenting llm-based logical reasoning from indeterminacy to determinacy},
  author={Sun, Hongda and Xu, Weikai and Liu, Wei and Luan, Jian and Wang, Bin and Shang, Shuo and Wen, Ji-Rong and Yan, Rui},
  booktitle={Proceedings of the 62nd Annual Meeting of the Association for Computational Linguistics (Volume 1: Long Papers)},
  pages={9828--9862},
  year={2024}
}

@inproceedings{sun2024harnessing,
  title={Harnessing multi-role capabilities of large language models for open-domain question answering},
  author={Sun, Hongda and Liu, Yuxuan and Wu, Chengwei and Yan, Haiyu and Tai, Cheng and Gao, Xin and Shang, Shuo and Yan, Rui},
  booktitle={Proceedings of the ACM Web Conference 2024},
  pages={4372--4382},
  year={2024}
}

@inproceedings{deng2025longdocurl,
  title={Longdocurl: a comprehensive multimodal long document benchmark integrating understanding, reasoning, and locating},
  author={Deng, Chao and Yuan, Jiale and Bu, Pi and Wang, Peijie and Li, Zhong-Zhi and Xu, Jian and Li, Xiao-Hui and Gao, Yuan and Song, Jun and Zheng, Bo and others},
  booktitle={Proceedings of the 63rd Annual Meeting of the Association for Computational Linguistics (Volume 1: Long Papers)},
  pages={1135--1159},
  year={2025}
}

@article{kang2026vlm,
  title={VLM-Loc: Localization in Point Cloud Maps via Vision-Language Models},
  author={Kang, Shuhao and Liao, Youqi and Wang, Peijie and Liao, Wenlong and Zhang, Qilin and Busam, Benjamin and Chen, Xieyuanli and Liu, Yun},
  journal={arXiv preprint arXiv:2603.09826},
  year={2026}
}

@article{lu2026mllms,
  title={Do MLLMs Really Understand Space? A Mathematical Reasoning Evaluation},
  author={Lu, Shuo and Cheng, Jianjie and Xu, Yinuo and Yu, Yongcan and Sheng, Lijun and Wang, Peijie and Jiang, Siru and Hu, Yongguan and Ling, Run and Shao, Yihua and others},
  journal={arXiv preprint arXiv:2602.11635},
  year={2026}
}

@article{zhang2025perl,
  title={Perl: Permutation-enhanced reinforcement learning for interleaved vision-language reasoning},
  author={Zhang, Yizhen and Ding, Yang and Zhang, Shuoshuo and Zhang, Xinchen and Li, Haoling and Li, Zhong-zhi and Wang, Peijie and Wu, Jie and Ji, Lei and Shen, Yelong and others},
  journal={arXiv preprint arXiv:2506.14907},
  year={2025}
}
